%% file: paper_template.tex
\documentclass[conference]{IEEEtran}
\usepackage{times}

\usepackage[numbers]{natbib}
\usepackage{multicol}
\usepackage[bookmarks=true]{hyperref}
\usepackage{amsmath}
\usepackage{amssymb}
\usepackage{graphicx}  
\usepackage{caption}    
\usepackage{cuted}
\usepackage{booktabs}
\usepackage{booktabs}
\usepackage{multirow}
\usepackage{xcolor}
\usepackage{subcaption}
\usepackage{marvosym}

\makeatletter
\newcommand{\rmnum}[1]{\romannumeral #1}
\newcommand{\Rmnum}[1]{\expandafter\@slowromancap\romannumeral #1@}
\makeatother

\begin{document}

\title{TwinRL: Digital Twin–Driven Reinforcement Learning for Real-World Robotic Manipulation}

\author{
\textbf{Qinwen Xu}$^{1,\star}$,
\textbf{Jiaming Liu}$^{1,\star,\dagger}$,
\textbf{Rui Zhou}$^{4,\star}$,
\textbf{Shaojun Shi}$^{1,\star}$,
\textbf{Nuowei Han}$^{1,\star}$,
\textbf{Zhuoyang Liu}$^{1}$,
\textbf{Chenyang Gu}$^{1}$,\\
\textbf{Shuo Gu}$^{2}$,
\textbf{Yang Yue}$^{3}$,
\textbf{Gao Huang}$^{3}$,
\textbf{Wenzhao Zheng}$^{3}$,
\textbf{Sirui Han}$^{4}$,
\textbf{Peng Jia}$^{2}$,
\textbf{Shanghang Zhang}$^{1,\textsuperscript{\Letter}}$
\\[0.5em]
$^{1}$State Key Laboratory of Multimedia Information Processing,
School of Computer Science, Peking University\\
$^{2}$Simplexity Robotics
$^{3}$Tsinghua University
$^{4}$Hong Kong University of Science and Technology
\\[0.5em]
\textbf{Project page:} \href{https://twinrl.github.io/}{https://twinrl.github.io}
\\[0.5em]
$^{\star}$Equal Contribution \quad
$^{\dagger}$Project lead \quad
$^{\textsuperscript{\Letter}}$Corresponding Author
    \vspace{-0.32cm}
}

\maketitle

\begin{strip}
    \centering
    \includegraphics[width=0.99\textwidth]{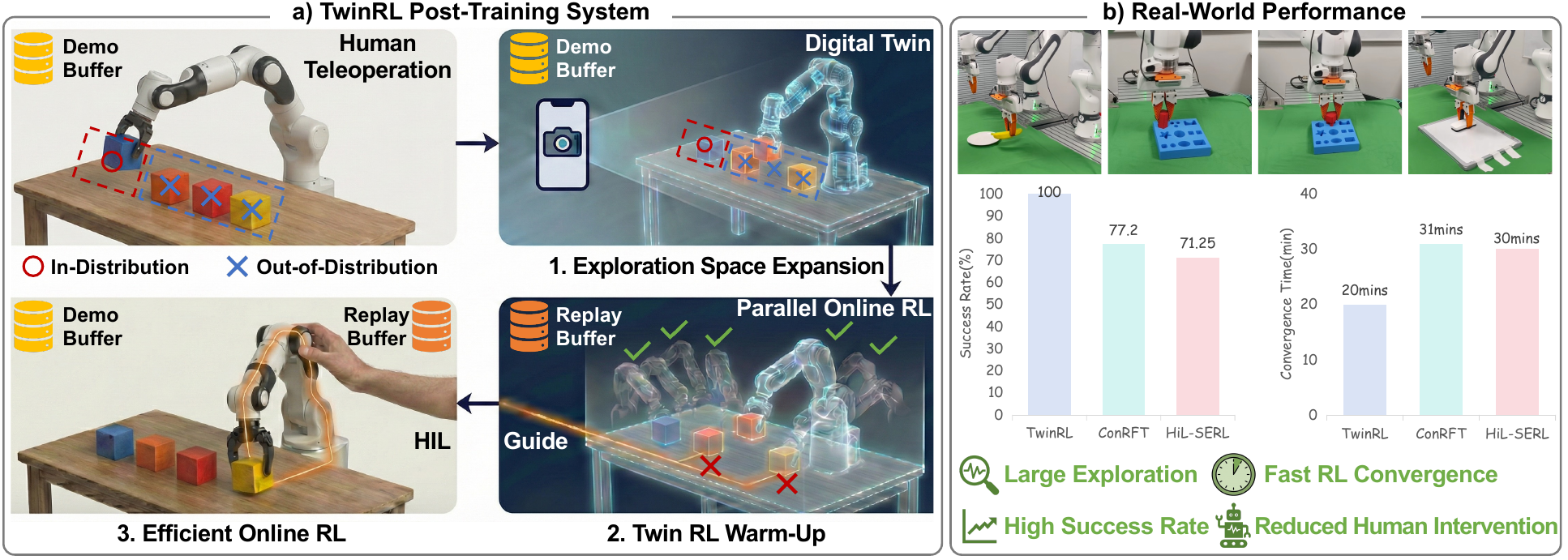}
    \vspace{-0.1cm}
    \captionof{figure}{\textbf{Overview.}
(a) To improve the efficiency and reliability of real-world RL, we propose \textbf{TwinRL}, a three-stage digital twin–real-world collaborative framework: (i) \textbf{SFT warm-up} expands the exploration space beyond real demonstrations; (ii) \textbf{twin RL warm-up} generates interactive trajectories to stabilize on-robot learning; and (iii) \textbf{real-world RL} leverages twin-identified failure-prone configurations to enable targeted HiL rollouts.
(b) Across four tasks, TwinRL converges faster and reaches near-100\% success in ~20 minutes across in- and out-of-distribution regions; HiL-SERL is evaluated only in-distribution.
}
    \label{fig:intro}
\end{strip}

\begin{abstract}
\input{section/abstract}

\end{abstract}
\vspace{-0.1cm}

\input{section/introduction}

\input{section/relatedwork}

\input{section/method}

\input{section/experiment}

\IEEEpeerreviewmaketitle

\section{Conclusion and Limitation} 
\label{sec:conclusion}

In this work, we identify a critical bottleneck in real-world VLA RL: the effective exploration space is largely shaped by the trajectory distribution induced during SFT, making adaptation in regions not covered by SFT data difficult even with HiL. To address this challenge, we present TwinRL, a digital twin–real-world collaborative framework that leverages digital twins not merely as simulators, but as exploration amplifiers and guides to accelerate real-world online RL. TwinRL combines (i) an exploration space expansion strategy that synthesizes diverse trajectories in the twin to broaden SFT coverage, and (ii) a twin RL warm-up strategy that uses parallel twin rollouts to initialize the real-world replay buffer and identify informative, failure-prone configurations for targeted HiL interventions. Experiments across four real-world manipulation tasks show that TwinRL substantially improves sample efficiency and robustness. 
TwinRL provides a practical and novel pathway for deploying online RL on physical robots. While it mitigates the offline-to-online transition gap, minor performance gaps may still arise in early stages; future work will further leverage informative twin RL trajectories to improve stability.

\bibliographystyle{plainnat}
\bibliography{references}

\clearpage

\begin{strip}
    \centering
    \includegraphics[width=0.99\textwidth]{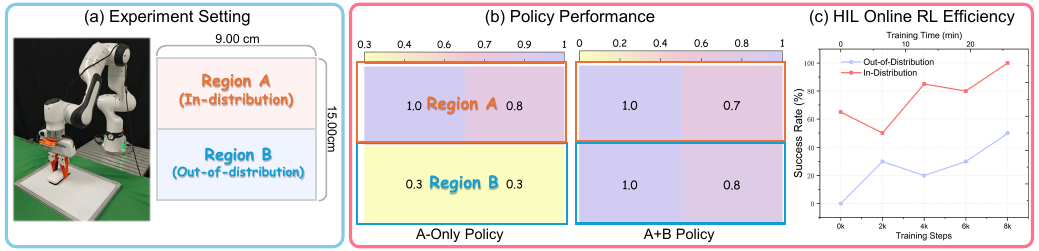}
    \vspace{-0.1cm}
    \captionof{figure}{\textbf{Motivation.} (a) We split the workspace into an in-distribution region (A) and an out-of-distribution region (B), defined by the manipulated object’s center location at task completion. (b) Heatmaps visualize the success rate of different policies. (c) Learning curves show the online RL training dynamics of the A-only policy in both regions.
}
    \label{fig:appendix_motivation}
\end{strip}

\appendix

\section*{A.\ Additional Motivation Experiments}
\label{sec:appendix_motivation}

Similar to the motivating observation in the main text, and motivated by the general-domain study~\cite{yue2025does}, we establish an analogous finding in robotic manipulation: exploration in real-world VLA RL is effectively constrained by the spatial support of the trajectory distribution induced during SFT. As shown in Fig.~\ref{fig:appendix_motivation} (a), beyond the main text, we further validate this phenomenon on the \textit{Erase-Whiteboard} task and identify a dual bottleneck: (1) it limits the set of states that can be reliably explored, and (2) it substantially reduces online RL efficiency, even with human intervention.

\label{sec:appendix_pick_banana_motivation}

\noindent\textbf{Experimental Setup.}
The experimental setup is identical to that in Section~III-B. We use the same criterion to split the workspace into an in-distribution Region~A and an out-of-distribution Region~B that is not observed during SFT. Below, we present the motivation results on the \textit{Erase-Whiteboard} task.

\noindent\textbf{Bottleneck~\Rmnum{1}.}
We compare two SFT data distributions with different spatial coverage: \textbf{(\rmnum{1}) A-only}, with 30 real-world demonstrations from Region~A, and \textbf{(\rmnum{2}) A+B}, which augments A-only with 30 digital-twin demonstrations from Region~B. We first evaluate the spatial generalization of the resulting SFT policies. We then initialize online RL from the A-only SFT policy and run autonomous training in the previously unseen Region~B to assess how demonstration coverage shapes the effective exploration space.

\noindent\textbf{Findings~\Rmnum{1}.}
As shown in Fig.~\ref{fig:appendix_motivation} (b), the success-rate heatmaps show a clear coverage-induced generalization gap. The \textbf{A-only} policy achieves high success within Region~A but fails almost everywhere in Region~B, indicating limited spatial extrapolation to uncovered configurations. In contrast, the \textbf{A+B} policy not only preserves strong performance in Region~A, but also substantially improves success in Region~B.
More importantly, running autonomous online RL in Region~B from the A-only initialization results in a clear exploration deadlock: when started from OOD configurations, the policy fails to reliably obtain positive rewards even after two hours of training.
These results corroborate that the effective exploration space is tightly bounded by the spatial support of the trajectory distribution induced during SFT: without Region~B coverage, the policy struggles to produce reward-yielding behaviors when initialized in OOD states, leading to exploration stall.

\noindent\textbf{Bottleneck~\Rmnum{2}.}
To alleviate the exploration deadlock, we follow prior real-world RL methods~\cite{luo2025precise, chen2025conrft} and use Human-in-the-Loop (HiL) intervention to guide the robot toward successful task completion. However, it remains unclear whether human guidance alone guarantees efficient online adaptation in OOD settings. To investigate this, we compare two regimes: In-distribution post-training, where online RL is performed in the familiar Region~A, and OOD post-training, where online RL is conducted in the unfamiliar Region~B. All models are initialized from the same A-only SFT policy, and HiL intervention is applied in both regimes.

\begin{figure*}[t]
    \centering
    \includegraphics[width=\textwidth]{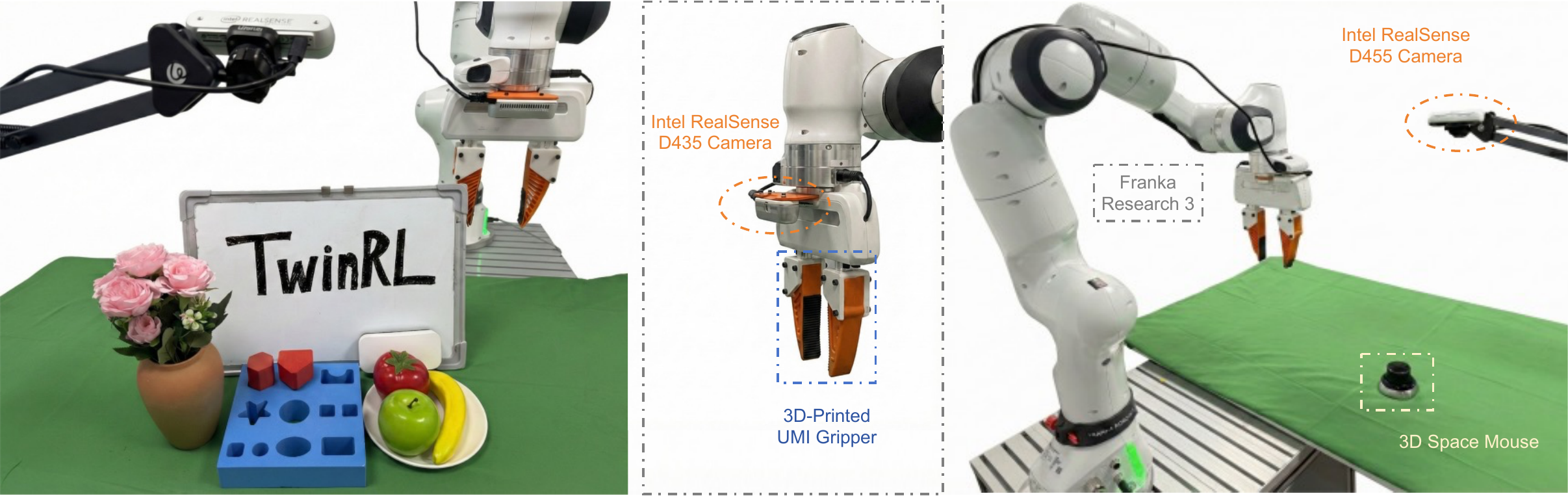}
    \caption{Real-World Robot Setups and Experimental Assets.}
    \label{fig:real-world-setup}
\end{figure*}

\noindent\textbf{Findings~\Rmnum{2}.}
The learning curves reveal a pronounced sample-efficiency disparity. In-distribution post-training improves rapidly and reaches a high success rate with limited interaction, whereas OOD post-training increases much more slowly and exhibits noticeably larger instability under the same budget, failing to match the in-distribution performance.
These results suggest that, even with HiL, learning in the previously unseen Region~B remains difficult because the replay buffer becomes highly imbalanced (across ID vs.\ OOD samples, teleoperation vs.\ RL trajectory styles, and successful vs.\ failed episodes), leading to inefficient and unstable gradient updates.

\noindent\textbf{Conclusion.}
Overall, the additional motivation experiments echo the dual exploration bottlenecks observed in Sec.~III-B:
(1) the spatial coverage that online RL can explore effectively is constrained by the spatial support of SFT data,
and (2) online adaptation in OOD settings is markedly less sample-efficient. These observations further motivate TwinRL to introduce digital twins to expand exploration coverage \emph{before} real-world interaction and systematically improve online learning efficiency.

\section*{B. Real-world Set-up}
\label{ap:RWS}
The real-world deployment of TwinRL is built on a modular robotic platform designed to support precise manipulation and efficient online reinforcement learning (RL) with reduced human intervention. All experiments are conducted on a 7-DoF Franka Emika Research 3 (FR3) robotic arm equipped with the 3D-printed UMI gripper~\cite{chi2024universal}, enabling accurate Cartesian end-effector control for contact-rich, precision manipulation tasks. As shown in Fig.~\ref{fig:real-world-setup}, we employ a dual-camera sensing setup to provide rich visual feedback for both perception and control. A fixed third-person RGB camera (Intel RealSense D455) captures a global view of the workspace, while a wrist-view RGB camera (Intel RealSense D435) is attached to the end effector to provide close-range observations. At each timestep, the observation consists of two RGB images: one from the wrist-view camera (resized to $128 \times 128$) and one from the third-person camera (resized to $256 \times 256$).

All real-world demonstrations and human-in-the-loop (HiL) interactions are collected via teleoperation using a 3D SpaceMouse, which enables operators to provide high-quality expert trajectories as well as targeted corrective interventions during online learning. To ensure a fair comparison, all methods are evaluated using identical hardware, sensing, and control software~\cite{schneider_franky}. The actor and learner processes for real-world online RL run on a workstation equipped with an NVIDIA GeForce RTX 4090 GPU, supporting policy inference and asynchronous training. The supervised fine-tuning (SFT) stage is performed offline on a server equipped with NVIDIA A100 GPUs (80GB).
\section*{C. Additional Details of the Digital Twin}
\label{ap:ADDT}
\begin{figure*}[htbp]
    \centering
    \includegraphics[width=0.95\textwidth]{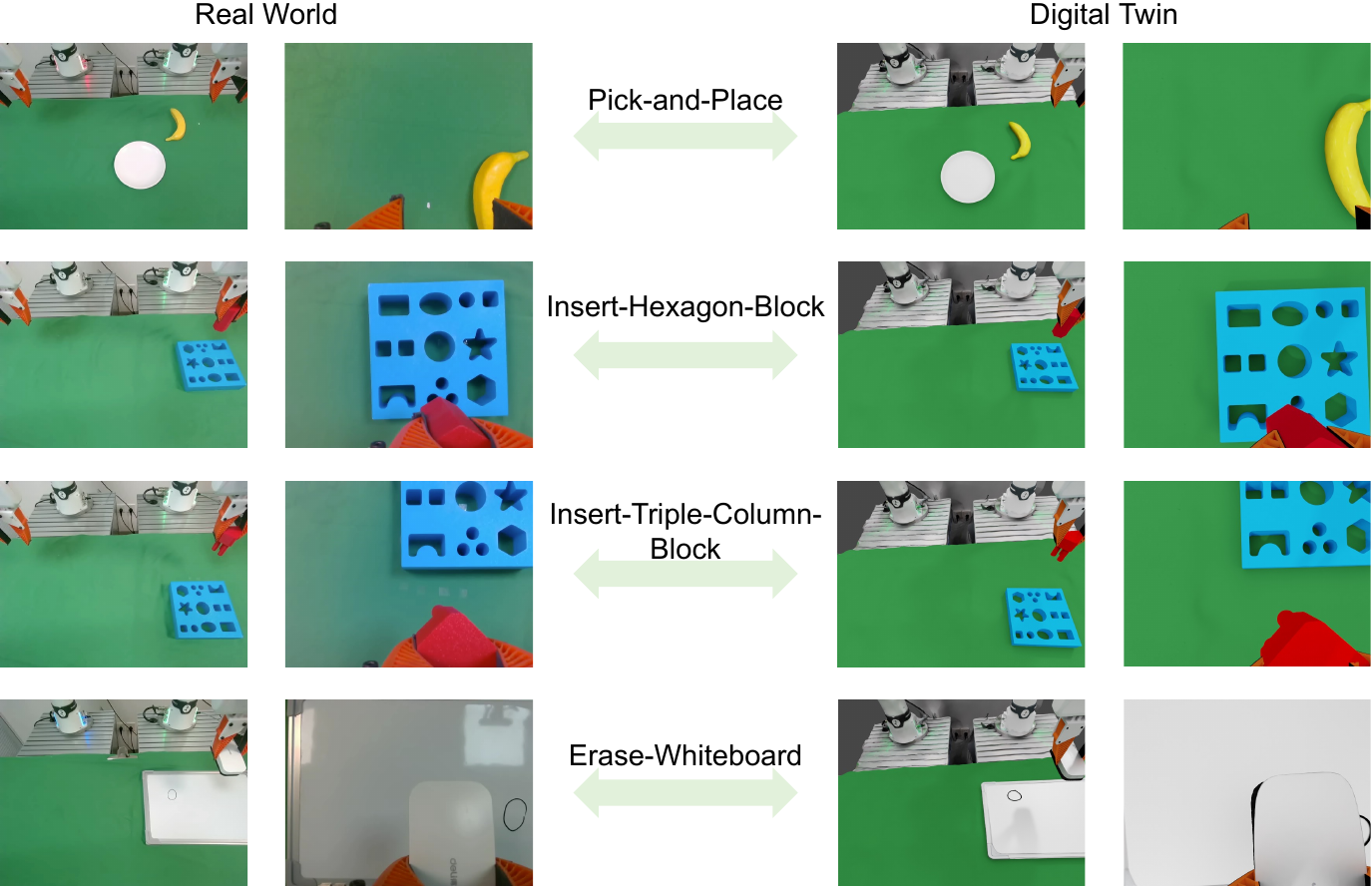}
    \caption{Qualitative comparison between real-world scenarios and their corresponding digital-twin renderings.}
    \label{fig:twin_vis}
\end{figure*}

In this section, we provide additional implementation details of our digital twin construction and object-centric modeling. The approximate time required for each stage is shown in Table~\ref{tab:twin_cost}.

\paragraph{\textbf{Digital Twin Construction.}}
The digital twin is instantiated from casually captured smartphone videos. Specifically, we record a single video of approximately one minute by moving a handheld smartphone around the target scene, ensuring full coverage of the robot workspace, including the complete robot base and surrounding environment. The video is captured at 30 FPS with a resolution of 1080×1920. The extracted frames are used to reconstruct the static scene geometry via 3D Gaussian Splatting (3DGS) tool~\cite{kiri_engine}. 
\begin{table}[t]
\centering
\caption{The time required to construct a digital twin scene for each task.}
\label{tab:twin_cost}
\renewcommand{\arraystretch}{1.15}
\setlength{\tabcolsep}{1pt}
\begin{tabular*}{\columnwidth}{@{\extracolsep{\fill}} c c}
\toprule
\textbf{Task} & \textbf{Time to Complete} \\
\midrule
Scanning & 1 min \\
Scene Reconstruction & 10 mins \\
Object Reconstruction & 5 s \\
Alignment & 3 mins \\
Grasp Pose Detection & 15 s \\
\bottomrule
\end{tabular*}
\end{table}

Following~\cite{yang2025noveldemonstrationgenerationgaussian}, to automate the alignment of the digital twin with the real-world setup, we anchor the coordinate transformation to URDF-defined robot frames.
 We first obtain a coarse initialization using point-cloud-based registration (e.g., ICP~\cite{besl1992sensor}). However, even after the coarse point-cloud-based alignment, residual misalignment may still remain, which can lead to inconsistencies between geometric alignment and visual appearance. 
To mitigate this issue, we re-implement the differentiable 3DGS rendering method and use it to refine the alignment.
Specifically, we select N camera viewpoints and, for each view, render a binary segmentation mask from the robot URDF model, denoted as $\mathcal{I}_i^{\text{URDF}}$, as well as a segmentation mask rendered from the robot’s 3D Gaussian representation under the current transformation, denoted as $\mathcal{I}_i^{\text{GS}}$. Pixels corresponding to the robot are assigned a value of 1, and all other pixels are set to 0. The alignment objective is defined as the mean pixel-wise discrepancy between the two masks across all viewpoints:
\begin{equation}
\mathcal{L}_{\text{align}} = \frac{1}{N}\sum_{i=1}^{N} \left(\mathcal{I}_i^{\text{URDF}} - \mathcal{I}_i^{\text{GS}}\right)^2.
\end{equation}
Starting from the ICP-based initialization, we parameterize the relative transformation—including translation, rotation, and scale—as $\hat{T}_{\text{rel}}$. Using differentiable rendering, the robot 3D Gaussian model $\hat{g}_{\text{robot}}$ is rendered from each camera viewpoint after applying $\hat{T}_{\text{rel}}$, yielding the following optimization objective:
\begin{equation}
\mathcal{I}_i^{\text{GS}}=\text{Render}(\hat{g}_{\text{robot}} \circ \hat{T}_{\text{rel}, \text{Camera}_i})^2.
\end{equation}
By exploiting the differentiability of 3DGS, we optimize $\hat{T}_{\text{rel}}$ via backpropagation and gradient descent, resulting in a refined alignment that enforces both geometric and visual consistency between the reconstructed robot and the real-world coordinate frame.

\paragraph{\textbf{Object-Centric Pose Estimation}}
We adopt an object-centric representation to model the manipulated object and its interaction with the robot. Given RGB observations, the target object is first reconstructed using SAM3D~\cite{chen2025sam}, producing a dense 3D representation of the object geometry. The reconstructed object is then uniformly subsampled to a fixed number (100k) of points to form an object-centric point cloud. 
We augment the object point cloud with a planar point cloud representing the supporting tabletop beneath the object. The combined point cloud is fed into AnyGrasp~\cite{fang2023anygrasp}, which returns grasp hypotheses ranked by confidence. We then select the top-n candidates, each represented as a 6-DoF grasp pose in the object-centric coordinate frame.
After initializing the object pose and its 6-DoF trajectory in the scene, we leverage the frame alignment results to transform these object-centric grasp poses into the robot coordinate system. Specifically, the estimated relative transformation between the reconstructed scene and the real-world robot base frame enables us to compute the time-varying end-effector poses in the robot base frame.
As a result, both the robot motion and the object trajectory are consistently defined within a shared coordinate frame in the digital twin. 

\paragraph{\textbf{Trajectory Generation}}
Following~\cite{yu2025real2render2real}, we adopt a collision-aware kinematic interaction model that prioritizes task-relevant visual–geometric consistency. Concretely, manipulated objects are treated as kinematic entities, and object–robot interactions are realized through frame-by-frame pose specification, together with simplified proxy collision geometries, non-penetration constraints, and workspace constraints.
This design preserves robot kinematics and basic geometric feasibility for trajectory synthesis and rollout generation. It also naturally aligns with the visuomotor policy setting considered in this work. Meanwhile, the simulation-to-real gap in physical modeling is inevitable, which further highlights the importance of Stage III in TwinRL-real-world online RL for policy refinement under physical execution. To generate diverse and task-consistent trajectories, we systematically vary the object’s initial configurations, target poses, and motion paths within the digital twin. Following prior work~\cite{shridhar2022peract}, for each task we extract a set of keyframes corresponding to critical stages of manipulation. Each keyframe specifies an object pose, from which the corresponding grasp pose and end-effector pose can be derived. We consider two complementary approaches for trajectory generation:

\noindent\textbf{Motion-Planning-Based Trajectory Generation.} Given the object poses at the keyframes, we directly employ motion planning toolkits~\cite{sucan2012the-open-motion-planning-library} to generate collision-free and kinematically feasible end-effector trajectories that connect the corresponding grasp poses. This approach ensures geometric consistency and feasibility under kinematic constraints.

\noindent{\textbf{Demonstration-Based Trajectory Augmentation.}
We leverage a single human teleoperated demonstration trajectory
$\tau = \{ \mathbf{x}_t \}_{t=0}^{T-1}, \quad \mathbf{x}_t \in \mathbb{R}^7$,
where T denotes the trajectory length, and each state $\mathbf{x}_t = (x_t, y_t, z_t, r_t, p_t, y_t, g_t)$ consists of the end-effector position, orientation (roll–pitch–yaw) and gripper state.
To synthesize new execution trajectories, we utilize the trajectory interpolation scheme from~\cite{yu2025real2render2real}. For the translational component, we compute an affine transformation that jointly aligns the start and end poses of the original teleoperated trajectory to the desired target start and end poses, and apply this transformation to the translational component of the entire trajectory. For the rotational component, we interpolate end-effector orientations along the trajectory using spherical linear interpolation, ensuring smooth and continuous rotation.

Finally, given the object and end-effector trajectories, the corresponding robot and object 3D assets are assembled in the digital twin through kinematic transformations and rendered to produce paired visual observations and robot states for subsequent policy learning. A qualitative comparison between real-world executions and their digital-twin renderings is shown in Fig.~\ref{fig:twin_vis}.

\section*{D. Additional Results}
\label{ap:AR}
\paragraph*{\textbf{D1. Episode Length}}

\begin{figure}[t]
\centering
\includegraphics[width=1.0\linewidth]{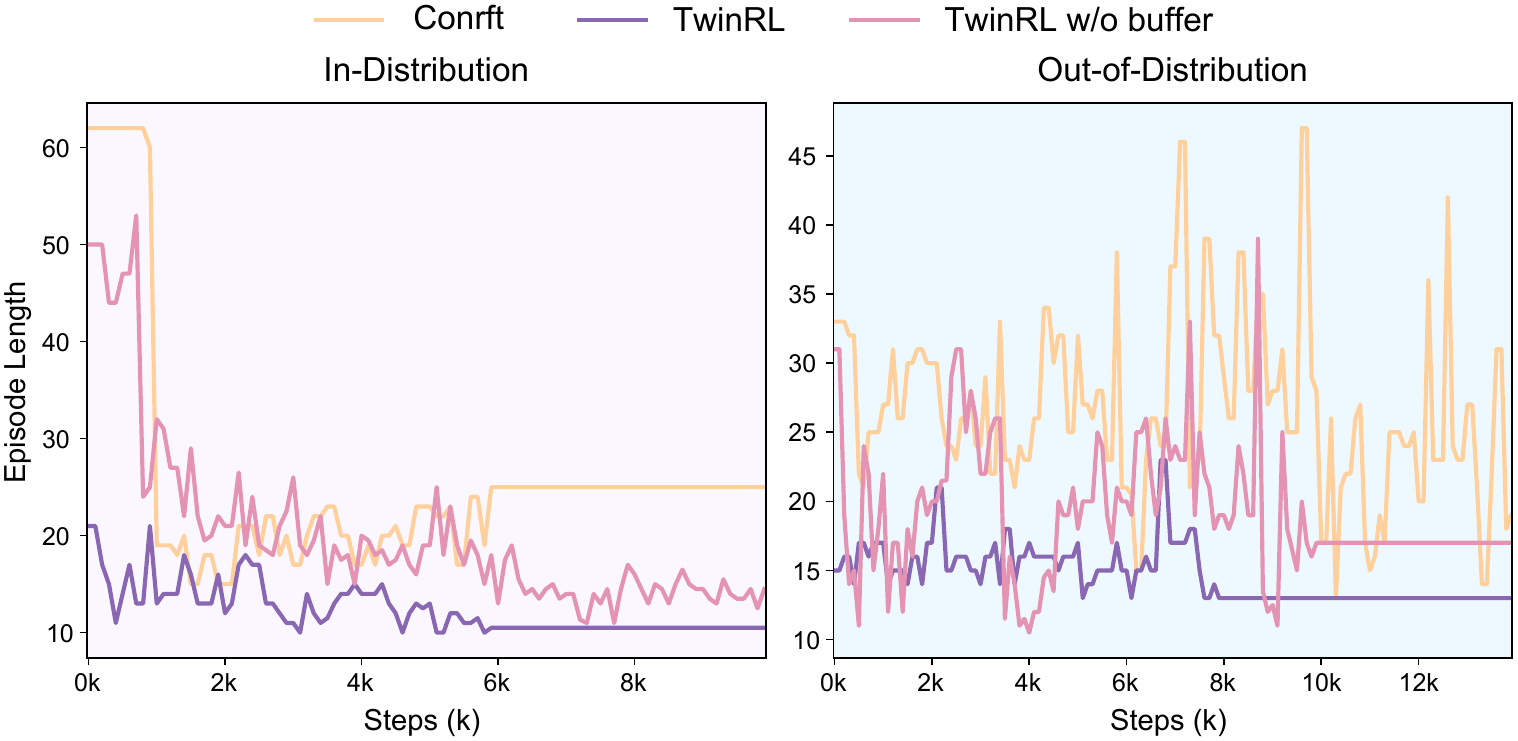}
\caption{\textbf{Episode length.} We report episode length curves for online RL in insert-triple-column-block
 manipulation tasks under both ID and OOD settings. The y-axis shows episode length, and the x-axis reports our model training steps.}
\label{fig:twinrl_episode_length}
\end{figure}

We provide additional analysis using episode length to characterize the efficiency of real-world online reinforcement learning.
All experiments are conducted on the Insert-Triple-Column-Block task under both in-distribution (ID) and out-of-distribution (OOD) settings, following the same real-world training protocol and baselines as in Section~IV-B of the main paper.
In this study, episode length is measured as the total number of low-level control steps executed within an episode, which terminates upon success, failure, or reaching a predefined step limit.
Fig.~\ref{fig:twinrl_episode_length} reports episode length as a function of real-world training time and training steps for all methods.
Across both ID and OOD regions, all methods exhibit a gradual reduction in episode length as training progresses, indicating that online reinforcement learning gradually leads to more direct and efficient execution patterns.
\textbf{In ID regions}, TwinRL demonstrates a markedly faster reduction in episode length, suggesting that twin-guided exploration provides more informative warm-up and early online rollouts, allowing the policy to focus on high-value state–action sequences and avoid redundant corrective motions.
\textbf{In OOD regions}, the difference becomes more pronounced. 
Baseline methods often require substantially longer episodes throughout training, indicating inefficient exploration and repeated recovery behaviors when encountering configurations outside the SFT data distribution.
Comparing TwinRL with TwinRL w/o buffer further highlights the role of the twin replay buffer.
While TwinRL w/o buffer already benefits from twin-guided exploration, TwinRL achieves shorter episode lengths earlier and with greater stability, indicating that initializing real-world training with twin RL rollouts helps bridge the offline-to-online transition and mitigates inefficient early exploration.

\paragraph*{\textbf{D2. Intervention Rate}}

\begin{figure}[t]
    \centering
    \includegraphics[width=\linewidth]{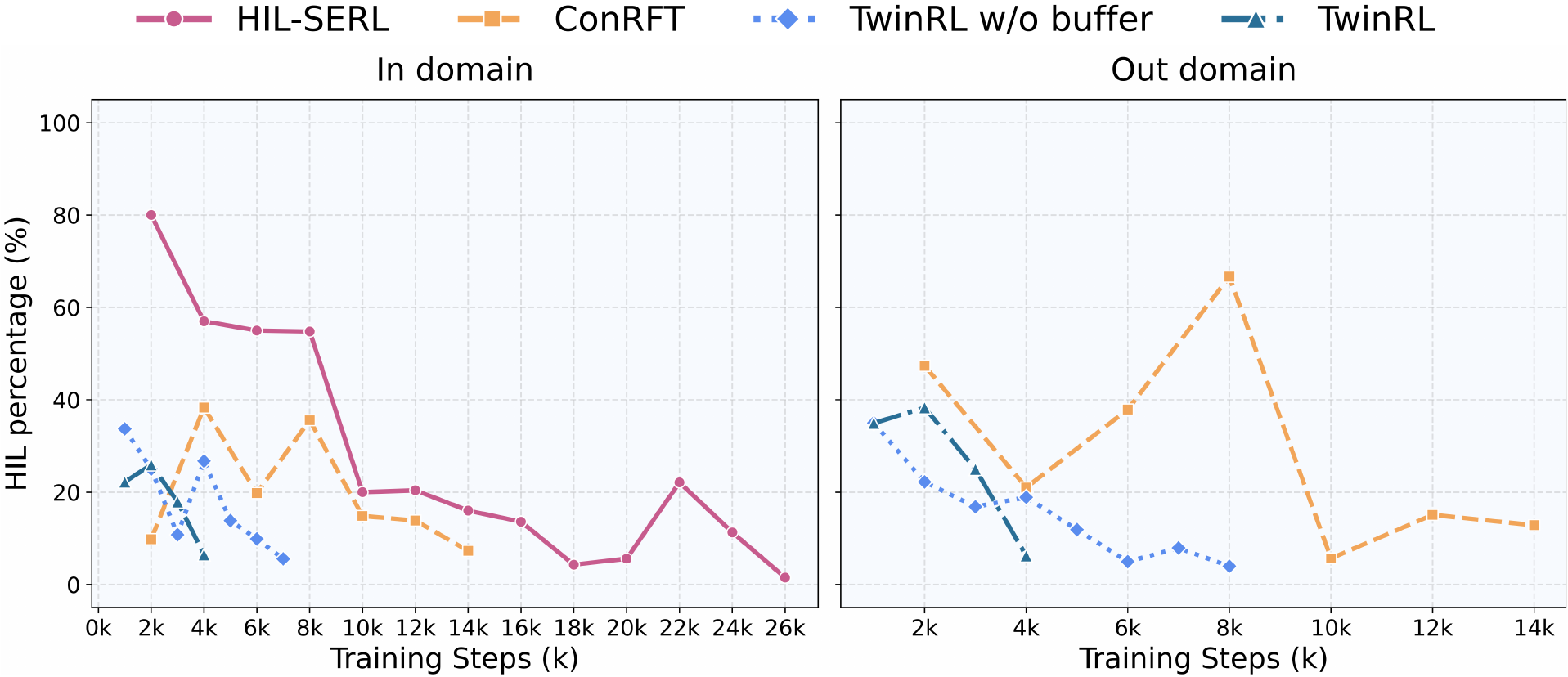}
    \caption{
        \textbf{Intervention Rate.}
        We report HiL intervention rate curves for online RL in Insert-Hexagon-Block manipulation tasks under both ID and OOD settings.
        The y-axis shows the HiL intervention percentage, and the x-axis reports our model training steps.
    }
    \label{fig:hil_rate}
\end{figure}

To further quantify improvements in human involvement efficiency during real-world online RL, we report the Human-in-the-Loop (HiL) intervention rate, defined as the fraction of steps within each rollout episode that require human assistance. We evaluate all methods on the Insert-Hexagon-Block task under both in-distribution (ID) and out-of-distribution (OOD) settings.
As shown in Fig.~\ref{fig:hil_rate}, HIL-SERL exhibits a high initial intervention rate in the ID setting (approximately 80\% at 2k steps), reflecting its strong reliance on continuous human intervention to bootstrap early-stage policy learning. Although the intervention rate gradually decreases, it remains relatively high over a substantial portion of the interaction process.
ConRFT starts with a moderate intervention rate, benefiting from its offline warm-up stage, but exhibits significant fluctuations during training. The situation further deteriorates in the OOD setting, where its intervention rate remains persistently high and unstable, highlighting the difficulty of adapting to unseen configurations without targeted guidance.
In contrast, TwinRL and TwinRL~w/o~buffer enter the real-world RL phase with lower initial intervention rates and converge to near-zero intervention much faster under both ID and OOD settings. This is because the TwinRL warm-up stage pre-populates the replay buffer with RL-style interaction trajectories, enabling a smoother offline-to-online transition. Meanwhile, the digital twin continuously identifies failure-prone yet informative object configurations, ensuring that each human intervention provides maximally informative signals for policy improvement rather than being triggered randomly.
These results demonstrate that, compared to prior methods, TwinRL substantially reduces the overall human intervention burden while achieving higher final success rates, validating the practical value of digital twins as exploration guides in real-world robotic RL.

\paragraph*{\textbf{D3. Additional Ablation Study}}

\begin{figure}[t]
    \centering
    \begin{subfigure}[t]{0.48\linewidth}
        \centering
        \includegraphics[width=\linewidth]{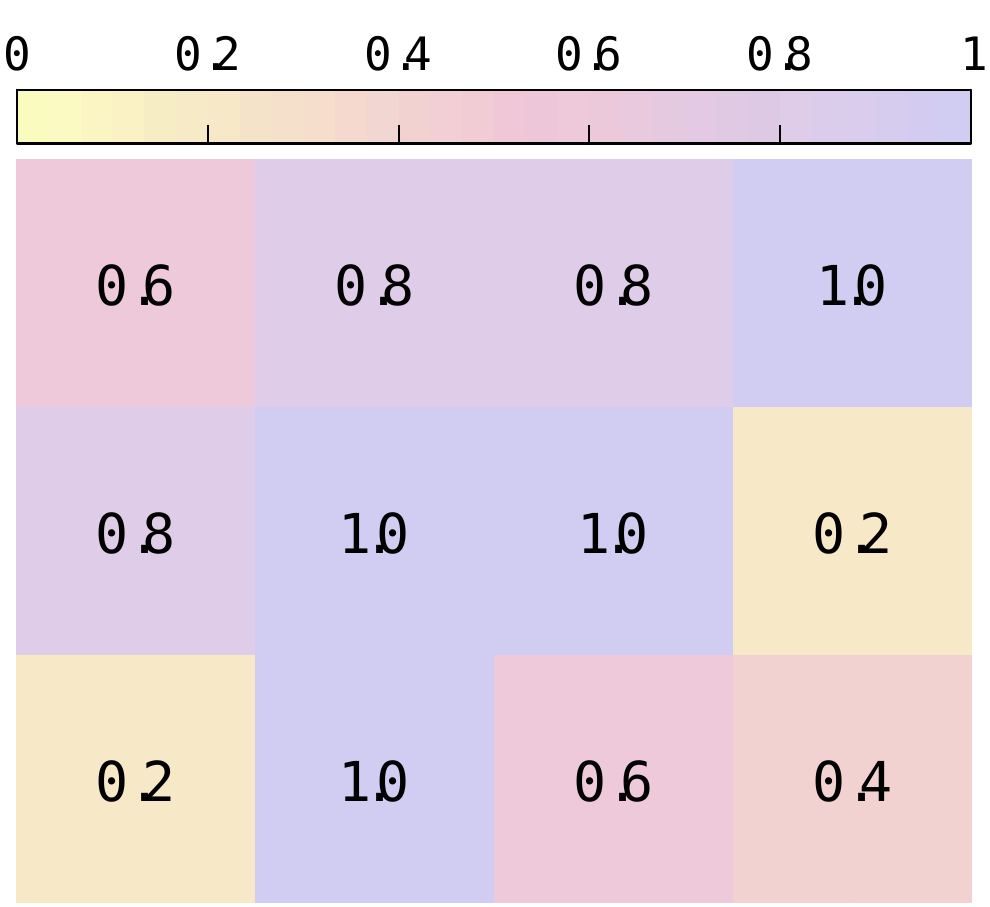}
        \caption{OOD-only augmentation}
        \label{fig:xxx_ood_only}
    \end{subfigure}
    \hfill
    \begin{subfigure}[t]{0.48\linewidth}
        \centering
        \includegraphics[width=\linewidth]{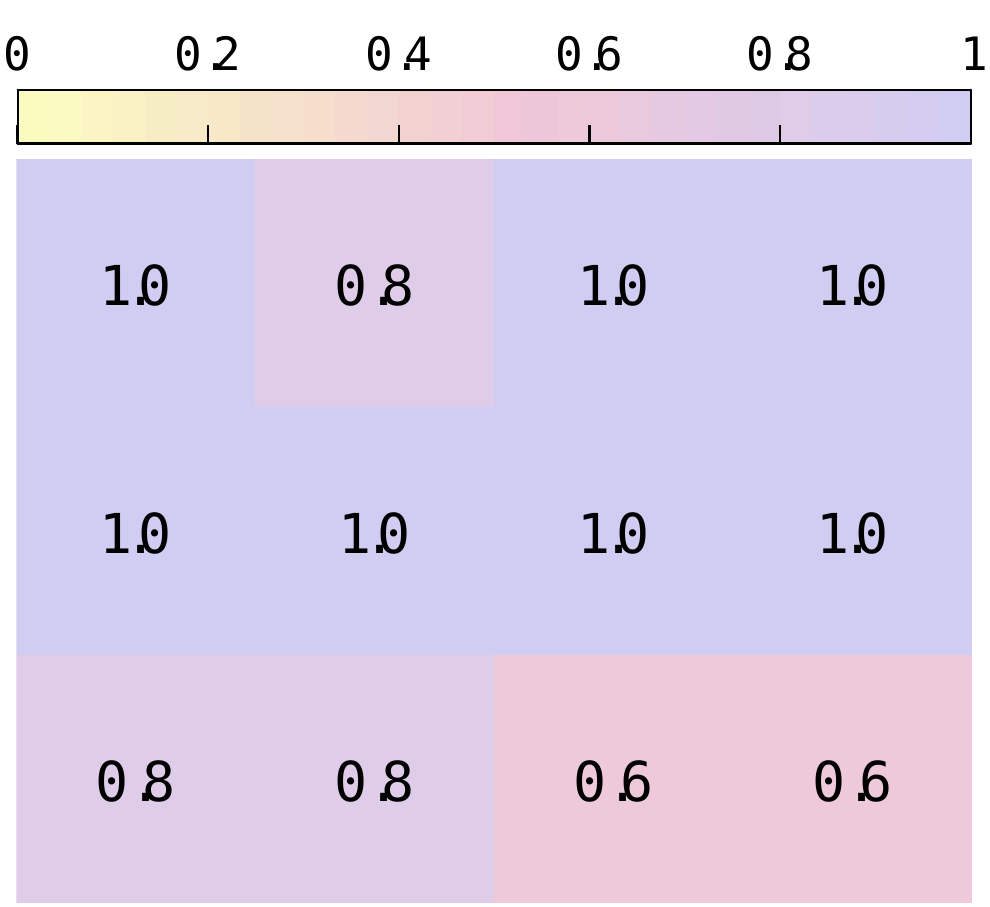}
        \caption{Joint augmentation}
        \label{fig:xxx_joint}
    \end{subfigure}
    \caption{
    \textbf{Additional ablations.} We evaluate how augmenting SFT with digital-twin data affects performance, comparing (a) adding twin data only in the OOD region and (b) adding twin data in both the ID and OOD regions. Each grid cell is evaluated with five rollout trials.
    }
    \label{fig:xxx}
\end{figure}

We conduct an additional ablation study on the Insert-Hexagon-Block task to further examine how the distribution of twin data affects warm-up SFT performance. 
In this study, we fix the amount of real-world data to 30 in-distribution (ID) trajectories and assess whether allocating synthetic twin trajectories within the ID region improves alignment between real-world and digital-twin domains.
Specifically, we compare two settings: OOD-only augmentation (ID 0 / OOD 30) and joint augmentation (ID 30 / OOD 30).
Fig.~\ref{fig:xxx} reports the corresponding success-rate heatmaps, where each grid cell is evaluated with five rollout trials.
Compared to OOD-only augmentation, jointly augmenting both ID and OOD regions leads to more consistent and uniformly higher success rates across the workspace.
While adding twin data in the OOD region already helps expand the exploration space and improves performance in previously uncovered areas, augmenting the ID region further improves performance, particularly within the in-distribution region.
These results suggest that, beyond expanding coverage, including twin data in the in-distribution region helps reduce the gap between synthetic and real data, yielding a more reliable warm-up policy.

\paragraph*{\textbf{D4. Evaluation in the Digital Twin}}
To quantitatively assess how well the digital twin captures the real-world task difficulty landscape, we conduct a controlled evaluation under matched initial configurations. This landscape fidelity is critical for TwinRL’s exploration space expansion and sim-to-real guided exploration, so we compare task success rates between real-world and digital-twin executions.
We focus on the Insert-Hexagon-Block task and evaluate the SFT checkpoint trained with mixed real-world and digital-twin data, before any online reinforcement learning is performed. 
The task workspace is discretized into in-distribution and out-of-distribution grids, identical to those used in the real-world evaluation. For each grid cell, the same policy is evaluated for 5 rollouts, both on the physical robot and in the corresponding digital-twin environment. 
Fig.~\ref{fig:twin_eval} reports the resulting success-rate heatmaps for real-world execution and digital twin execution. The results show that the digital twin accurately reflects the global structure of task difficulty observed in the real world. 
Although absolute success rates in the digital twin may differ from those in the real world, the relative ordering of easy versus difficult regions is well preserved. As a result, sampling and prioritizing these configurations during twin-based rollouts effectively identifies informative states for subsequent real-world exploration.
These results demonstrate that digital twins in TwinRL function not merely as simulators, but as effective exploration amplifiers and guides, enabling systematic identification of challenging configurations and improving the efficiency of real-world reinforcement learning.
\begin{figure}[t]
    \centering
    \begin{subfigure}[t]{0.48\linewidth}
        \centering
        \includegraphics[width=\linewidth]{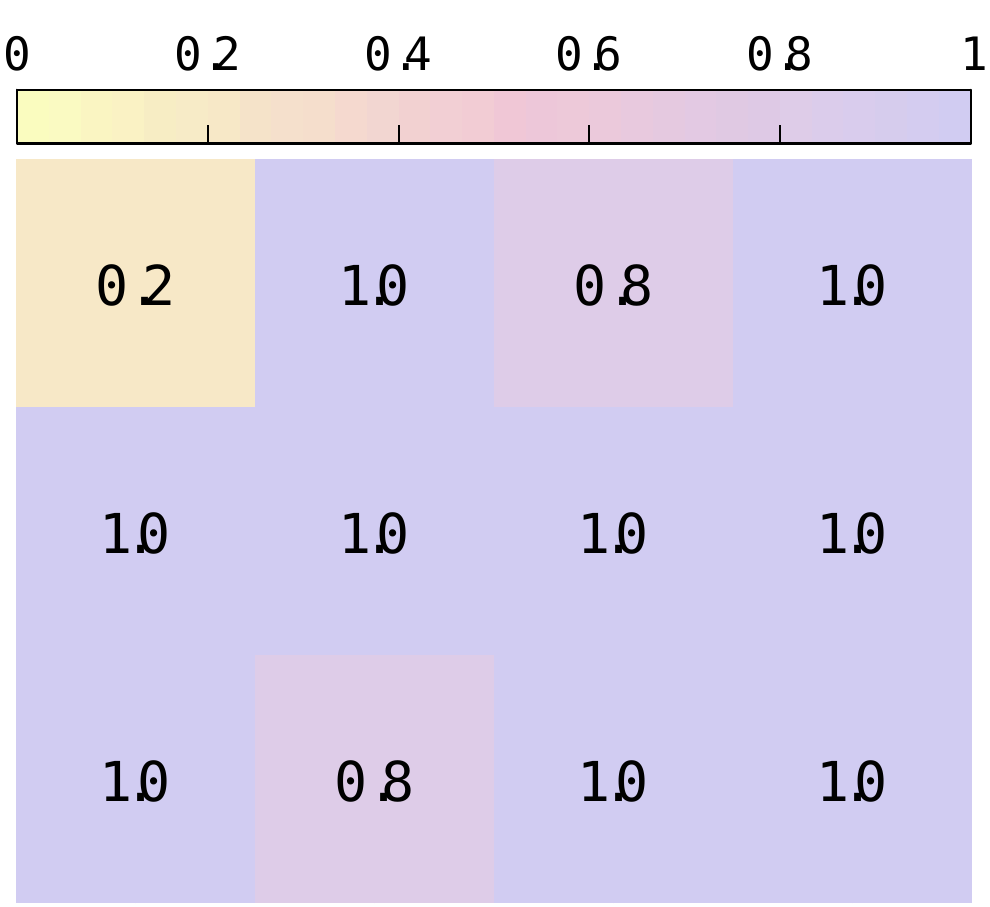}
        \caption{Real world}
        \label{fig:xxx_Real}
    \end{subfigure}
    \hfill
    \begin{subfigure}[t]{0.48\linewidth}
        \centering
        \includegraphics[width=\linewidth]{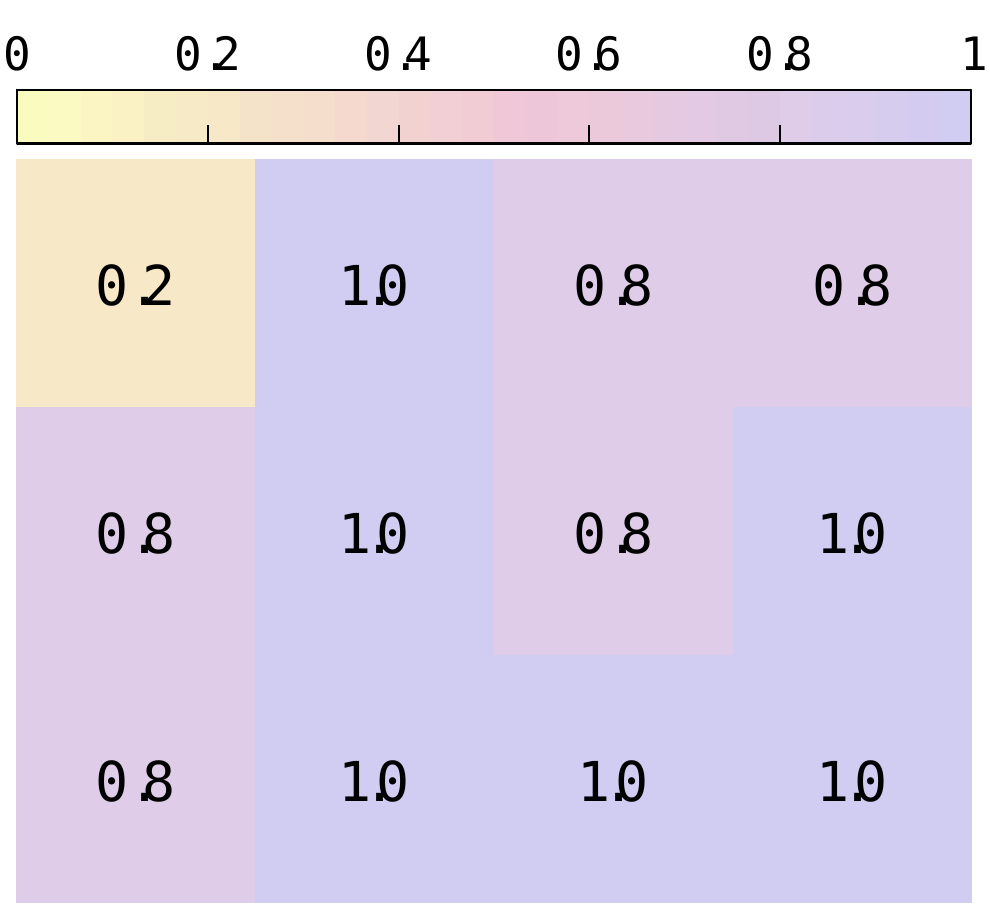}
        \caption{Digital Twin}
        \label{fig:xxx_Twin}
    \end{subfigure}
    \caption{Comparison of success-rate heatmaps between the real-world scenario and the corresponding digital-twin scenario using the same SFT model on the Insert-Hexagon-Block task.}
    \label{fig:twin_eval}
\end{figure}

\begin{figure}[t]
\centering
\includegraphics[width=1.0\linewidth]{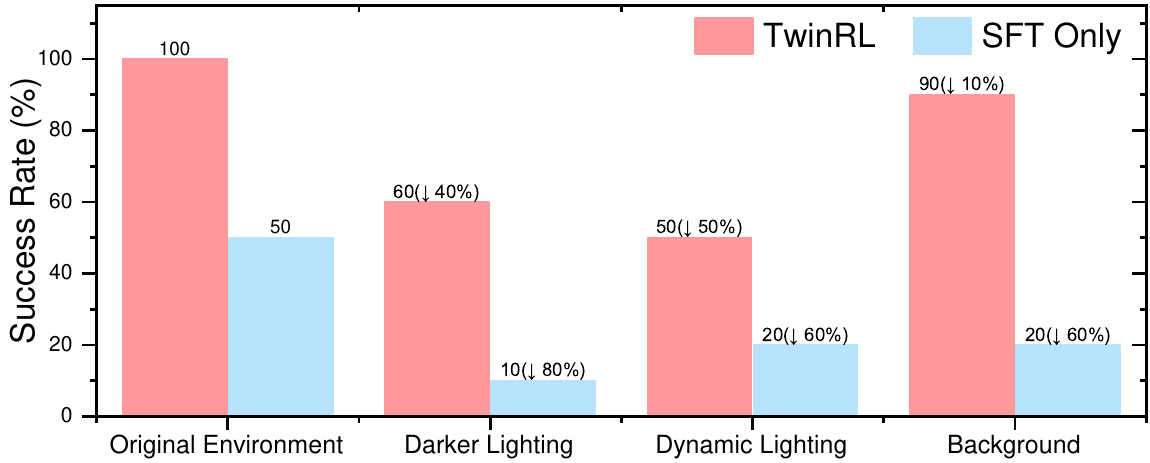}
\caption{\textbf{Additional Robustness Evaluation.} We additionally compare the SFT policy and the TwinRL-guided online RL policy on the Pick-and-Place task under previously unseen environmental perturbations.}
\label{fig:pick_banana_robust_generalization}
\end{figure}

\begin{figure*}[t]
    \centering
    \includegraphics[width=\textwidth]{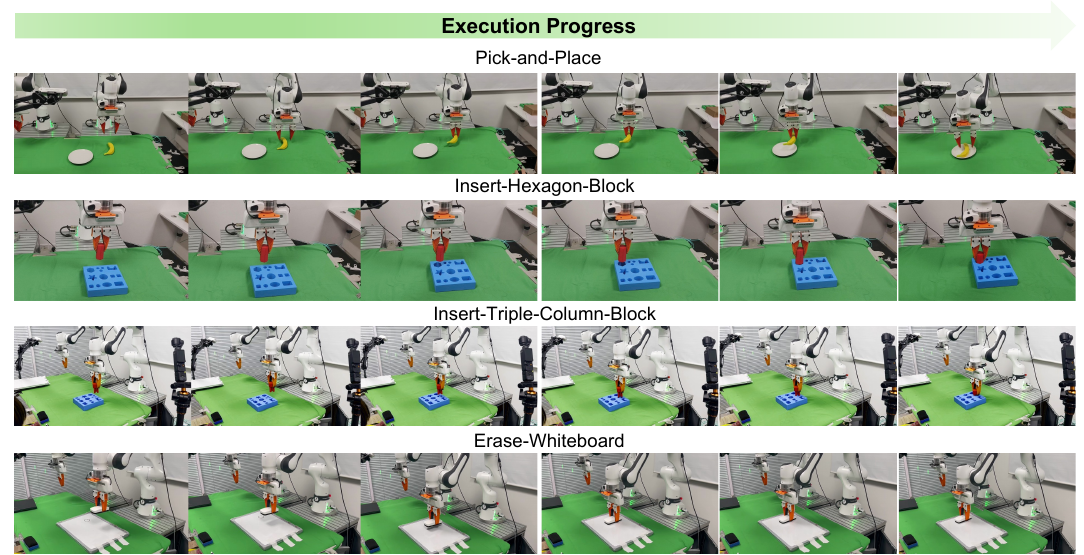}
    \caption{Visualization of complete real-world task execution processes (left to right).}
    \label{fig:success_cases}
\end{figure*}

\paragraph*{\textbf{D5. Additional Robustness Evaluation}}
To further evaluate our approach's generalization capability beyond the insertion task, we extend the zero-shot robustness evaluation to the Pick-and-Place task. Consistent with the analysis in Section~IV-D, we compared the SFT policy and the TwinRL-guided policy in three categories of previously unseen environmental perturbations: Background (introduction of distractors and clutter), Darker lighting (low-light conditions), and Dynamic lighting (changing colored light patterns).
As shown in Fig.~\ref{fig:pick_banana_robust_generalization}, the results of the Pick-and-Place task align with our previous findings. The SFT baseline shows significant performance degradation under distribution shifts, particularly during the grasping phase, where background clutter causes visual ambiguity. In contrast, the TwinRL policy demonstrates remarkable resilience, maintaining a high success rate with only a slight decrease in performance compared to the original environment.
These results further corroborate that TwinRL does not merely memorize visual features from demonstrations. Rather, the twin-guided HiL mechanism effectively directs the policy to explore and recover from high-risk states. This fosters the learning of robust, noise-tolerant decision boundaries, which are critical for robotic manipulation.

\begin{figure*}[t]
    \centering
    \includegraphics[width=\textwidth]{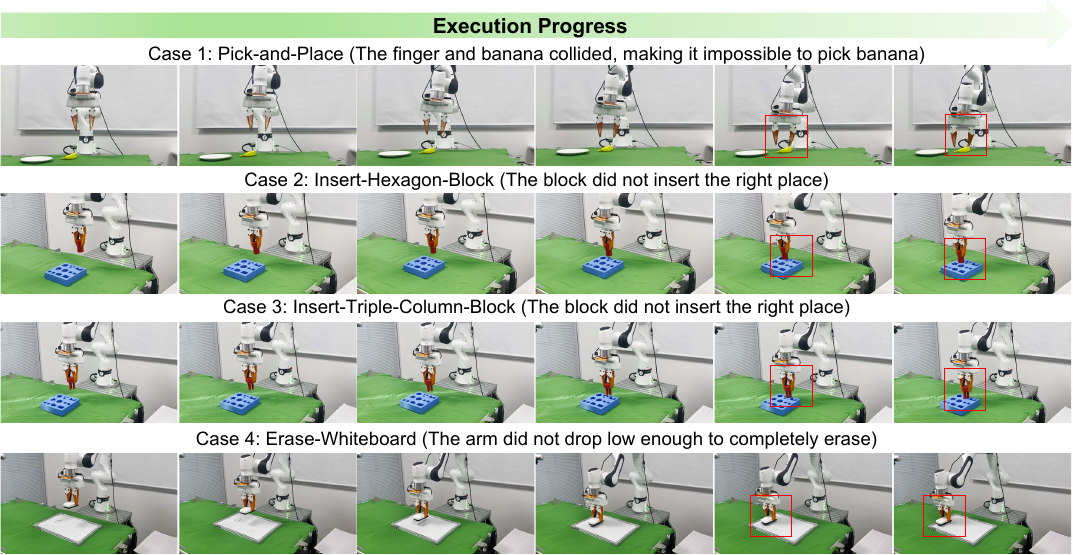}
    \caption{Visualization of failure cases on different tasks, and the red box highlights the failure positions.}
    \label{fig:failure_cases}
\end{figure*}
\paragraph*{\textbf{D6.Real-World Visualizations}} 
Fig.~\ref{fig:success_cases} illustrates representative task executions across all four of our tasks, including Pick-and-Place, Insert-Hexagon-Block, Insert-Triple-Column-Block, and Erase-Whiteboard in single-arm settings. We observe that TwinRL produces smooth and continuous motions, especially for precise actions such as insertion.

\section*{E. Failure Case Analysis}
\label{ap:FCA}

In the four tasks designed for this study, several typical failure modes were identified. As shown in Figure~\ref{fig:failure_cases}, the main objects and regions where issues occurred are highlighted with red boxes, and the causes of failure are analyzed below.

\paragraph*{\textbf{Imprecise Object Detection}} 
In the Pick-and-Place task, the model is required to accurately grasp a banana from various positions on the table and place it onto a plate. However, during this process, the robotic arm collides with the banana as it descends, causing the object to shift and resulting in a failure to grasp. This issue is particularly prominent in the data augmentation regions, but improves as the online reinforcement learning (RL) progresses.

\paragraph*{\textbf{Positional Instability}}
In the Insertion tasks (Task 2 and Task 3), the robotic arm is required to insert a block with precision into a designated slot on the board, with the error constrained to within 2 millimeters. However, given the extensive task scope and stringent precision requirements, the robotic arm may encounter friction with the board while reaching certain distant positions. This friction results in the block's loss of proper alignment, either in terms of angle or position, thereby preventing it from achieving the correct placement. Consequently, variations in insertion angle and position inevitably arise, culminating in task failure.

\paragraph*{\textbf{Manipulation height}}
In the Erase-Whiteboard task, the FR3 robotic arm is assigned the task of utilizing an eraser to remove a predetermined pattern from a whiteboard. However, the robotic arm sometimes fails to reach the required height for effective erasure, resulting in improper alignment of the eraser with the whiteboard surface and incomplete erasure. Moreover, the absence of visual feedback during task execution hinders the system's capacity to ascertain in real-time whether the pattern has been fully erased, thereby elevating the probability of failure.

\end{document}

%% file: section/abstract.tex
Despite strong generalization capabilities, Vision-Language-Action (VLA) models remain constrained by the high cost of expert demonstrations and limited real-world interaction. While online reinforcement learning (RL) has shown promise, its application to real-world VLA manipulation is hindered by low exploration efficiency and restricted exploration coverage. Through systematic real-world experiments, we observe that the effective exploration space of online RL is largely constrained by the trajectory distribution induced during supervised fine-tuning (SFT). Motivated by this observation, we propose \textbf{TwinRL}, a digital twin–real-world collaborative post-training framework that expands and guides RL exploration for VLA models through three stages: SFT warm-up, twin RL warm-up, and real-world RL. TwinRL first reconstructs a high-fidelity digital twin from smartphone-captured scenes. During the SFT stage, we introduce an exploration space expansion strategy that expands the support of the trajectory distribution beyond real demonstrations, reshaping the exploration space for more effective RL. Rather than treating the twin as a data augmentation tool, we propose a twin RL warm-up strategy that enables it to act as an exploration guide for real-world RL. Specifically, TwinRL performs efficient parallel RL in the digital twin to generate interactive trajectories that populate the replay buffer and stabilize subsequent real-world RL learning. This process also identifies failure-prone yet informative configurations, enabling targeted human-in-the-loop rollouts to further improve on-robot efficiency.
Across four tasks, TwinRL achieves
near-100\% success in both in-distribution and out-of-distribution regions, delivering over 30\% faster convergence than prior real-world RL methods with only 20 minutes of on-robot interaction.

%% file: section/introduction.tex
\section{Introduction}
\label{sec:intro}

Building on internet-scale pretrained Vision–Language Models (VLMs)~\cite{alayrac2022flamingo,karamcheti2024prismatic,wang2024qwen2}, Vision–Language–Action (VLA) models have recently emerged as a promising paradigm for robotic manipulation~\cite{rt22023arxiv,kim2024openvla}. 
Through training on large-scale robot demonstrations, VLA models have achieved encouraging progress in robotic scene reasoning~\cite{lin2025onetwovla,liu2024robomamba}, generalization~\cite{intelligence2025pi05visionlanguageactionmodelopenworld,liu2025hybridvla}, and precise manipulation~\cite{bjorck2025gr00t,black2024pi_0}.
Despite these advances, current VLA models remain constrained by their heavy reliance on expensive expert demonstrations and the limited amount of real-world interaction, which restricts robustness in complex physical environments.

Reinforcement learning (RL) provides an exploration-based framework that reweighs cumulative task rewards, and has proven highly effective in enhancing the reasoning capabilities of general VLMs~\cite{schulman2017proximal, rafailov2023direct, shao2024deepseekmath}.
Motivated by this progress, recent studies have explored applying RL as a post-training stage for VLA models, leveraging offline RL updates~\cite{zhang2024grape,huang2025co,zhang2025reinbot} or online exploration in simulation~\cite{liu2025can,tan2025interactive,lu2025vla,guo2025improving} to refine manipulation policies.
While these approaches demonstrate that RL can improve execution accuracy, applying online RL to real-world manipulation remains limited.

First, unlike simulation environments, real-world robots must operate under safety-constrained, sequential interactions with physical objects, where parallel experimentation is infeasible, leading to severely limited efficiency for online learning~\cite{chen2025conrft}. While existing approaches improve efficiency through warm-up strategies~\cite{lei2025rl, li2025simplevla}, unified training objectives~\cite{chen2025conrft}, or human-in-the-loop (HiL) feedback~\cite{luo2025precise}, they remain strongly dependent on the quality, diversity, and timing of teleoperated demonstrations, often resulting in unstable convergence.
Second, through systematic real-world experiments, we identify a fundamental limitation of real-world RL for VLA models: the effective exploration space is tightly constrained by the trajectory distribution induced during supervised fine-tuning (SFT). Even with HiL assistance, learning in out-of-distribution (OOD) regions remains challenging due to an unfavorable reward landscape and an imbalanced replay buffer. This observation, consistent with findings in general domains~\cite{yue2025does}, suggests that RL primarily reweights existing behaviors rather than expanding beyond them. This insight highlights the importance of expanding the exploration space of VLA models for efficient post-training.

To address this challenge, we propose TwinRL, a digital twin–real-world collaborative post-training framework that leverages digital twins as exploration amplifiers and guides under real-world interaction constraints. 
Specifically, the overall pipeline consists of an SFT warm-up stage, followed by a digital RL warm-up and real-world RL stage.
During SFT, we introduce an exploration space expansion strategy that expands the support of the trajectory distribution by generating diverse synthetic trajectories within the digital twin. 
For in-distribution regions, paired real and synthetic samples are collected under matched configurations to facilitate alignment between real and simulated domains. For OOD regions, object configurations are uniformly sampled to systematically promote balanced coverage and enable exploration beyond the bias of demonstrated states.
To instantiate digital twins in novel environments, we reconstruct high-fidelity scenes from smartphone videos using 3D Gaussian Splatting and convert them into mesh representations~\cite{yu2025real2render2real}. Within the twin, we estimate object-centric 6-DoF poses~\cite{fang2023anygrasp}, providing a geometric interaction representation for generating realistic manipulation trajectories.

Building on this enriched initialization, we propose a twin RL warm-up strategy that enables the digital twin to act as an exploration guide for real-world RL. Rather than serving solely as a data augmentation tool, TwinRL integrates digital twin interaction and real-world feedback into a unified post-training pipeline. Specifically, TwinRL performs efficient parallel RL rollouts in the digital twin to generate interactive trajectories, which are used to enrich the replay buffer for real-world RL. This mitigates the distribution mismatch between SFT data and RL interactions, enabling a more stable transition to online learning while preserving catastrophic forgetting of high-precision behaviors.
While twin-based RL enables scalable exploration, real-world interaction remains necessary to account for physical dynamics. 
During real-world RL stage, TwinRL leverages extensive twin rollouts to identify failure-prone yet informative configurations, using them to guide targeted HiL rollouts. This avoids inefficient random exploration in the real world and significantly improves sample efficiency.
As shown in Fig.~\ref{fig:intro}b), we evaluate TwinRL on four manipulation tasks using a shared VLA backbone~\cite{octo_2023}. During the SFT stage, our exploration space expansion strategy improves average success by 42\% compared to training on real-world demonstrations alone. During online RL, TwinRL achieves near-100\% success in both in-distribution regions and spatial OOD settings, while reducing real-world interaction time by at least 30\% compared to prior methods~\cite{chen2025conrft, luo2025precise}. On average, TwinRL requires only about 20 minutes of on-robot interaction across four tasks. Beyond convergence speed and accuracy, we further demonstrate that TwinRL improves robustness under previously unseen conditions, including background clutter and lighting variations.
In summary, our contributions are as follows:

\begin{itemize}
    \item We observe that exploration in real-world VLA RL is largely shaped by the trajectory distribution induced during SFT. Based on this insight, we propose \textbf{TwinRL}, a digital twin–real-world collaborative post-training framework that leverages digital twins as exploration amplifiers and guides.

    \item We introduce an exploration space expansion strategy that generates diverse trajectories within the digital twin to broaden exploration coverage, together with an efficient pipeline for constructing high-fidelity digital twins.

    \item We propose a twin RL warm-up strategy that integrates twin interaction with real-world RL: twin rollouts provide RL-style experience to stabilize early-stage learning, while failure-prone configurations guide targeted HiL exploration on real robots.

\end{itemize}

%% file: section/relatedwork.tex
\section{Related Work}
\noindent\textbf{Vision-Language-Action (VLA)} models~\cite{kim2024openvla,black2024pi_0,li2024cogact,qu2025spatialvla,liu2025hybridvla,wen2025tinyvla,chen2025fast,wen2024diffusion,liu2026last0latentspatiotemporalchainofthought} ground high-level language instructions into visuomotor control, enabling robots to accomplish tasks in dynamic environments.
Recent works couple VLM representations with specialized action experts, such as diffusion-based heads~\cite{liu2024rdt,wen2024diffusion,li2024cogact,bjorck2025gr00t,ze20243d,jia2025video2act,gu2025manualvlaunifiedvlamodel} that generate actions via iterative denoising, or flow-based formulations~\cite{black2024pi_0,intelligence2025pi05visionlanguageactionmodelopenworld,su2025freqpolicy}.
While VLA policies are trained via supervised fine-tuning (SFT), their reliance on static demonstrations limits exploration beyond observed states.

\noindent\textbf{Reinforcement Learning (RL) for VLA Models.}
RL policies can continue to improve robustness post-deployment through online interaction and feedback from the environment~\cite{tan2025interactive}.
Existing RL post-training approaches for VLAs follow two primary paradigms. Offline methods~\cite{zhang2024grape,zhang2025reinbot} optimize policies using fixed trajectories or preference data to enhance alignment and robustness. Alternatively, online approaches~\cite{lu2025vla,tan2025interactive,chen2025tgrpo,guo2025improving,chen2025conrft} leverage interactive simulator rollouts and iterative updates to improve task performance and stability.
Recent work has started to bring \emph{real-world} RL to physical robots by training policies directly through on-robot interaction, often leveraging human-in-the-loop feedback and safety-aware data collection to improve sample efficiency\cite{luo2025precise,chen2025conrft,luo2023rlif,mendonca2024continuously,lei2025rl100performantroboticmanipulation,luo2024rlifinteractiveimitationlearning}. However, they typically require extensive real-world interaction and prolonged training, which are costly, low-throughput, and further constrained by safety risks.
Unlike prior methods, we systematically identify how the SFT trajectory distribution constrains effective exploration in real-world RL, and leverage a high-fidelity digital twin to expand the exploration space and improve efficiency during joint simulation and real-world post-training.

\noindent\textbf{Digital Twin-Based Sim-to-Real Data Scaling.}
To generate more training data from limited real demonstrations while mitigating the sim-to-real gap, many works~\cite{ye2025video2policy,lou2024robogsphysicsconsistentspatialtemporal,yang2025noveldemonstrationgenerationgaussian,xu2025r2rgenrealtoreal3ddata,xue2025demogen} adopt a Real-to-Sim-to-Real paradigm built upon a digital twin. DexMimicGen~\cite{jiang2025dexmimicgen} supports demonstration-based policy learning by organizing assets in simulation. RoboVerse~\cite{geng2025roboverse} provides larger-scale simulated interactions to promote manipulation. RialTo~\cite{torne2024reconciling} and CASHER~\cite{torne2024robot} integrate real observations with interactive simulation components. RoboGSim~\cite{li2025robogsimreal2sim2realroboticgaussian} combines 3D Gaussian Splatting with a physics engine to enable scalable data generation and strong sim-to-real transfer. Real2Render2Real~\cite{yu2025real2render2real} uses a phone scan and one human video to render robot training demos via 3D Gaussian Splatting. Additionally, GSWorld~\cite{jiang2025gsworld} enables closed-loop, photo-realistic simulation by combining 3D Gaussian Splatting with physics for reproducible evaluation and sim-to-real policy learning. Beyond using a digital twin in the SFT stage, SGFT~\cite{yin2025rapidly} learns a value function in simulation and uses it to reshape rewards and guide efficient real-world fine-tuning. SLAC~\cite{hu2025slac} learns a task-agnostic latent action space in simulation and applies it to downstream real-world RL to improve safety.
SimLauncher~\cite{wu2025simlauncher} uses a digital-twin–pretrained policy to bootstrap critic learning with simulated and real-world demonstrations, and to provide action proposals during RL.
Unlike prior work, we do not treat the digital twin merely as a data augmentation tool, but instead leverage it as an exploration amplifier and guide for efficient real-world RL.

%% file: section/method.tex
\section{Method}

\begin{figure*}[t]
\centering
\includegraphics[width=1.0\textwidth]{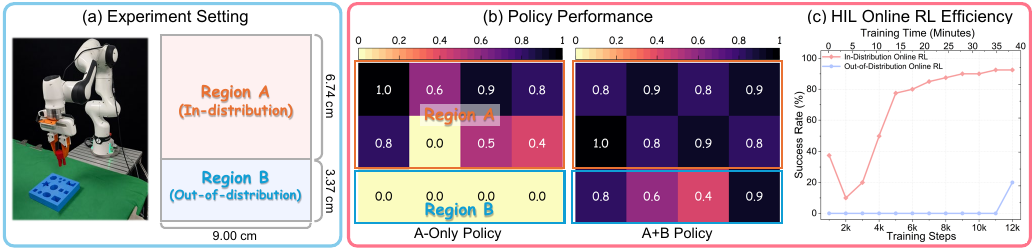}
\vspace{-0.3cm}
\caption{\textbf{Exploration Bottlenecks.} (a) We split the workspace into an in-distribution region (A) and an OOD region (B). Each region is defined by the manipulated object’s center location at task completion. (b) Heatmaps visualize the performance of different policies. (c) Learning curves show the online RL training dynamics of the A-only policy in both regions.}
\vspace{-0.3cm}
\label{fig:motivation}
\end{figure*}

\subsection{Preliminary}

\noindent\textbf{Vision-Language-Action (VLA) Policy Action Generation.} The VLA policy $\pi_\theta$ maps a language instruction $\ell$ and multi-view images $I_t=\{I_t^{\text{side}}, I_t^{\text{wrist}}\}$ to a 7-DoF end-effector action $a_t \sim \pi_\theta(a_t | I_t, \ell)$ at each timestep $t$.
The action $a_t = (\Delta p_t, \Delta r_t, g_t)$ defines the relative movement of the end-effector, comprising a 3D translational delta $\Delta p_t \in \mathbb{R}^3$, a 3D rotational change $\Delta r_t \in \mathbb{R}^3$, and a binary gripper state $g_t \in \{0,1\}$.We denote a rollout trajectory as $\tau = \{(I_t, \ell, a_t)\}_{t=1}^{T}$, which terminates upon task success or reaching a time limit.

\noindent\textbf{Reinforcement Policy.} Following previous works~\cite{chen2025conrft,luo2025precise}, reinforcement learning (RL) serves as an interactive post-training paradigm that leverages environmental feedback to optimize fine-grained manipulation and broaden state coverage through exploration. Robotic RL can be formulated as a Markov decision process (MDP) $\mathcal{M}=\{\mathcal{S},\mathcal{A},\rho,\mathcal{P},r,\gamma\}$, where $s\in\mathcal{S}$ denotes the state observation, $a\in\mathcal{A}$ denotes the action, $\rho(s_0)$ is the initial-state distribution, $\mathcal{P}$ is the unknown and potentially stochastic transition dynamics, $r:\mathcal{S}\times\mathcal{A}\rightarrow\mathbb{R}$ is the reward function, and $\gamma\in(0,1]$ is the discount factor. Under this setting, to evaluate the performance of a policy $\pi$, we define the state-value function as
\begin{equation}
V^{\pi}(s)=\mathbb{E}_{\pi}\Big[\sum_{t=0}^{H}\gamma^{t}r(s_t,a_t)\ \big|\ s_0=s\Big],
\end{equation}
and the action-value function is defined as
\begin{equation}
Q^{\pi}(s,a)=\mathbb{E}_{\pi}\Big[\sum_{t=0}^{H}\gamma^{t}r(s_t,a_t)\ \big|\ s_0=s,\ a_0=a\Big],
\end{equation}
which satisfy $V^{\pi}(s)=\mathbb{E}_{a\sim \pi(\cdot|s)}[Q^{\pi}(s,a)]$. We define the optimal policy $\pi^*$ as the one maximizing the expected return over trajectories $\tau$ sampled from the distribution of ($\rho, \mathcal{P}$, $\pi$):
\begin{equation}
\pi^* = \arg\max_{\pi} \mathbb{E}_{\pi} \left[ \sum_{t=0}^{H} \gamma^t r(s_t, a_t) \right].
\end{equation}
$\pi_\theta(a|s)$ is parameterized by a neural network and can be modeled as a Gaussian distribution for continuous control.

\subsection{Motivation}
\label{sec:motivation}

While online RL provides a pathway toward task robustness, its sample efficiency on physical hardware remains a challenge. Through systematic real-world experiments, we find that the trajectory space effectively explored by real-world VLA RL is tightly constrained by the spatial support induced during SFT. This constraint introduces a dual bottleneck: (1) it limits the set of states that can be reliably explored, and (2) it substantially reduces online RL efficiency, even with human intervention.
These motivation experiments reveal a key insight: expanding the underlying trajectory distribution is fundamental to improving RL exploration efficiency.

\noindent\textbf{Experimental Setup.} In Fig.~\ref{fig:motivation} (a), we conduct all experiments using a precision block insertion task that requires high positional accuracy. All policies are instantiated based on the Octo~\cite{team2024octo} model. We partition the workspace into an in-distribution Region A, which is covered by demonstrations, and an out-of-distribution (OOD) Region B that is not observed during SFT.

\noindent\textbf{Bottleneck~\Rmnum{1}.} The spatial coverage of SFT demonstrations is varied to isolate its impact on policy generalization and autonomous online RL.
Specifically, we compare two training data distributions: \textit{(\rmnum{1}) A-only}, with 30 demonstrations from Region A, and \textit{(\rmnum{2}) A+B}, which adds 30 digital-twin demonstrations from Region B.
We first evaluate the spatial generalization of the resulting SFT policies. To measure how demonstration coverage shapes the effective exploration space, we initialize the policy with the A-only SFT model and run autonomous online RL in the unseen Region B.

\noindent\textbf{Findings~\Rmnum{1}.} 
As shown in Fig.~\ref{fig:motivation}(b), we perform 10 rollouts within each grid cell.
In Regoin B, the \textit{A+B} policy achieves a success rate of 62.5\%, whereas the \textit{A-only} policy remains confined to Region~A (0\% in Region B). This result indicates that standard SFT policies exhibit limited extrapolation to spatially uncovered regions.
More importantly, attempting autonomous online RL in Region~B from the \textit{A-only} model leads to a pronounced exploration deadlock. When initialized in OOD configurations, the policy fails to consistently obtain positive rewards, even after 40K training steps (approximately two hours). 
These results show that, consistent with findings in the general-domain work~\cite{yue2025does}, the effective exploration space of online RL is closely tied to the spatial coverage of the SFT data.

\noindent\textbf{Bottleneck~\Rmnum{2}.}
To mitigate the exploration deadlock, Human-in-the-Loop (HiL) intervention can be introduced to guide the robot toward successful task completion~\cite{luo2025precise}. However, an important question remains: does the availability of human guidance guarantee efficient online adaptation in OOD settings? 
To examine this, we compare two settings: \textit{In-distribution Post-training}, where online RL is performed in the familiar Region~A, and \textit{OOD Post-training}, where online RL is conducted in the unfamiliar Region~B. All models are initialized from the same A-only SFT policy.

\noindent\textbf{Findings~\Rmnum{2}.} 
Despite human interventions providing successful corrective demonstrations in both settings, we observe a pronounced disparity in sample efficiency. As shown in Fig.~\ref{fig:motivation} (c), in-distribution post-training adapts rapidly, achieving over $90\%$ success within approximately 45 minutes ($\sim$14k interaction steps). In contrast, OOD post-training converges substantially more slowly and exhibits greater instability under the same interaction budget, failing to reach comparable performance. 
These results indicate that, even with the incorporation of a HiL scheme, learning in the previously unseen Region~B remains challenging due to an unfavorable reward landscape and an imbalanced data distribution in the data buffer, both of which substantially reduce gradient efficiency.

\noindent\textbf{Conclusion.} These observations suggest that overcoming both bottlenecks requires expanding exploration coverage \emph{before} real-world interaction and guiding human intervention to systematically improve online efficiency, rather than relying on random HiL. Motivated by this, we propose TwinRL, a digital twin–real-world collaborative RL framework that leverages digital twins as exploration amplifiers and guides for online RL. Additional motivation experiments in other tasks are provided in~\hyperref[sec:appendix_motivation]{Appendix~A}.

\begin{figure*}[t]
\centering
\includegraphics[width=0.99\textwidth]{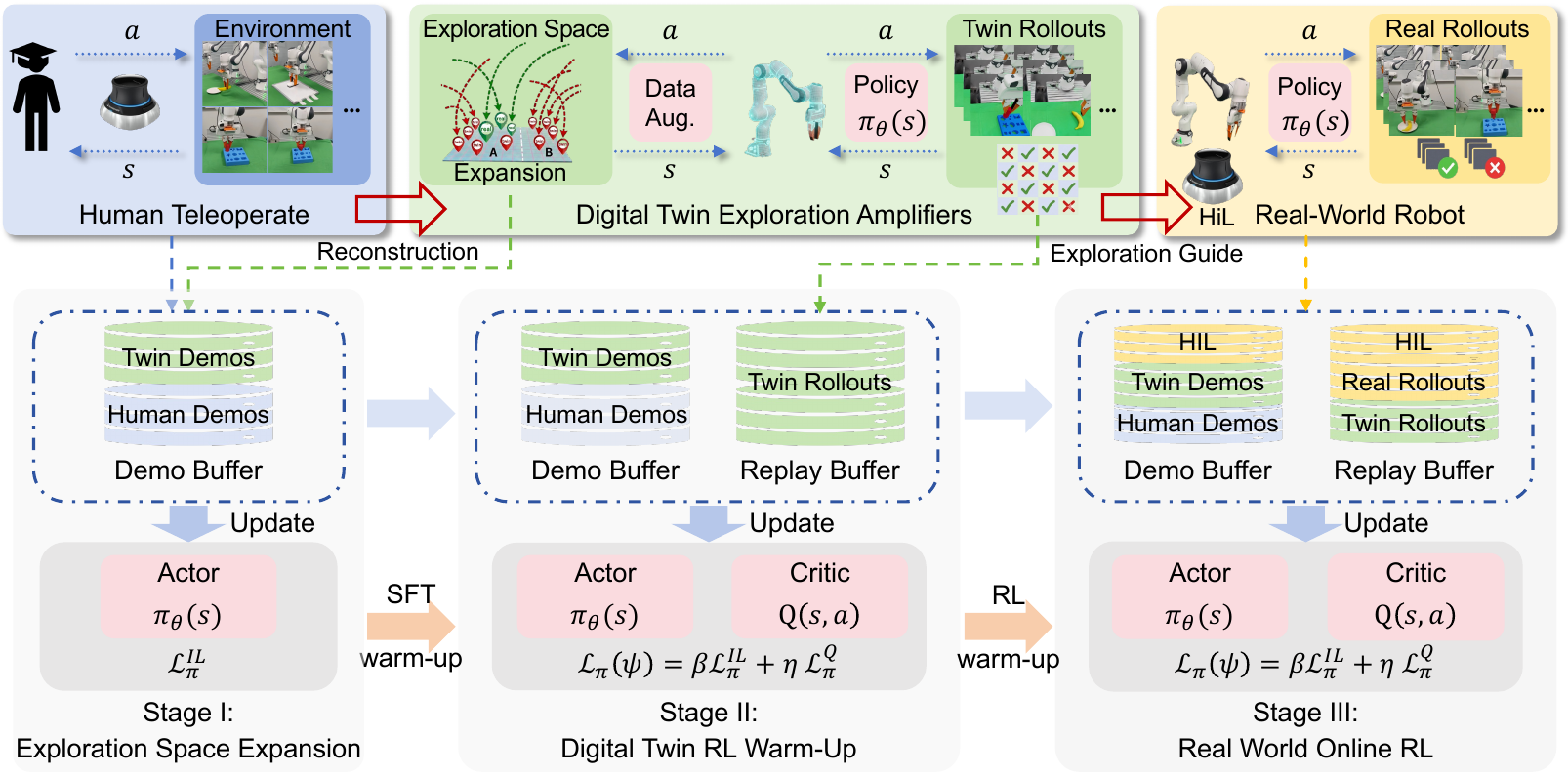}
\vspace{-0.1cm}
\caption{\textbf{TwinRL}. Stage I: Starting from human teleoperation, we introduce an exploration-space expansion strategy that synthesizes diverse digital-twin demonstrations to broaden SFT coverage. Stage II: The SFT-initialized policy is then trained with scalable, parallel online RL in the digital twin to harvest RL-style rollouts, which are transferred to initialize the real-world replay buffer and stabilize online learning. Stage III: During real-world online RL, the digital twin efficiently and continuously identifies failure-prone yet informative object configurations, which are then used to guide targeted HiL rollouts.}
\vspace{-0.3cm}
\label{fig:method}
\end{figure*}

\subsection{Exploration Space Expansion Strategy}
\noindent\textbf{Digital Twin Construction.}
We construct a high-fidelity digital twin of the target manipulation environment as the primary substrate for exploration expansion and virtual RL rollouts in TwinRL.
It is rapidly instantiated from casually captured smartphone videos by reconstructing the scene using 3DGS tools~\cite{kiri_engine} (about 10 minutes), the manipulable objects using SAM3D~\cite{chen2025sam} (about 5 seconds), and the robot from its URDF model. All components are unified as mesh-based assets for kinematic assembly and integrated into Blender or simulation environments~\cite{todorov2012mujoco}.
To enable bidirectional knowledge transfer, we align the digital twin with the real environment at both the visual and robot-state levels. Alignment is anchored to URDF-defined robot frames: we obtain a coarse match via point-cloud registration (e.g., ICP~\cite{besl1992sensor}) and refine it with differentiable 3DGS rendering~\cite{yang2025noveldemonstrationgenerationgaussian} to align rendered and real observations, yielding consistent coordinate frames.
To accelerate reconstruction, we adopt a collision-aware kinematic interaction model in the digital twin that prioritizes task-relevant visual–geometric consistency~\cite{yu2025real2render2real}. 
Manipulated objects are modeled as kinematic entities, and object–robot interactions are defined through frame-by-frame pose specification, together with simplified proxy collision geometries and workspace constraints.
To enable trajectory synthesis under this model, we adopt an object-centric representation by estimating the manipulated object’s 6-DoF grasp pose with AnyGrasp~\cite{fang2023anygrasp}, which defines the relationship between the object and the end effector. Starting from a single successful object trajectory, we then generate diverse execution trajectories via inverse kinematics, motion planning, or affine transformations. Empirically, this design provides effective data for expanding the exploration space and offers reliable guidance for real-world RL. More details are provided in~\hyperref[ap:ADDT]{Appendix~C}, including qualitative visual comparisons and a summary of manual effort and runtime overhead.

\noindent\textbf{Exploration Amplifiers.}
During the warm-up stage, we treat the digital twin as an \emph{exploration amplifier}, enriching trajectory diversity to expand the effective exploration space for subsequent online RL. As shown in Fig.~\ref{fig:method} Stage I, we generate diverse object-centric synthetic trajectories that go beyond real demonstrations by varying the object’s initial configurations, target poses, and motion paths. Given an object’s initial pose $T_0 \in \mathrm{SE}(3)$ and a desired target pose $T_{\text{target}} \in \mathrm{SE}(3)$, both randomly sampled with positional and orientation variations, we estimate a task-consistent grasp pose $T_{\text{grasp}}$ and derive the boundary end-effector poses:
\begin{equation}
T^{\text{ee}}_{\text{start}} = T_0 \cdot T_{\text{grasp}}, \quad T^{\text{ee}}_{\text{end}} = T_{\text{target}} \cdot T_{\text{grasp}}.
\end{equation}
The intermediate trajectory is generated via motion planning or by applying affine transformations to a single demonstration trajectory, ensuring trajectory quality.
Note that for a 30-step task, with parallel processing, we only need about 1 minute to construct a set of digital-twin demonstrations.
To internalize these augmented behaviors, we perform an SFT stage on the merged buffer $\mathcal{D}$ by minimizing the imitation learning loss:
\begin{equation}
\mathcal{L}_{\pi}^{\text{IL}} = -\mathbb{E}{(s,a) \sim \mathcal{D}} [\log \pi_{\psi}(a|s)].
\end{equation}
We use the digital twin not only to mitigate exploration deadlock in OOD regions, but also to collect additional in-distribution data, thereby narrowing the sim-to-real gap.

\begin{figure*}[t]
\centering
\includegraphics[width=0.99\textwidth]{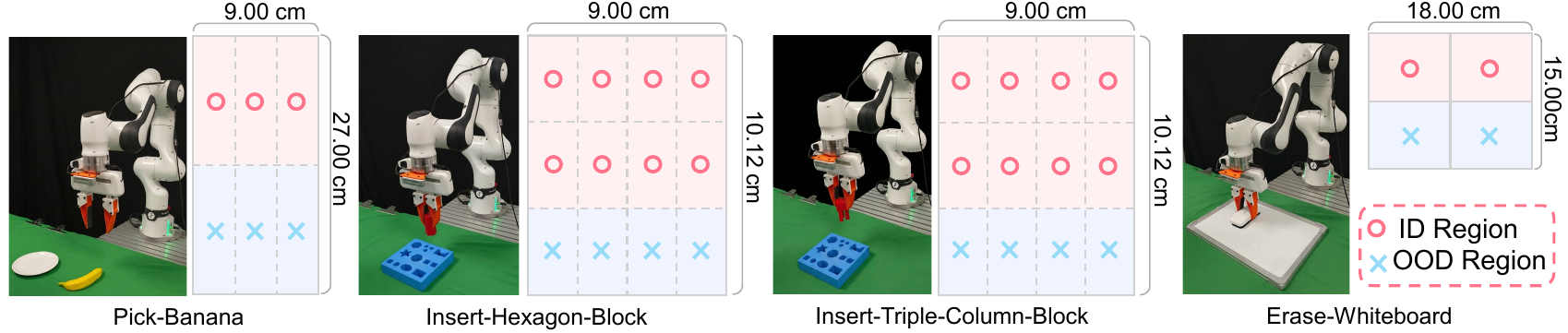}
\vspace{-0.1cm}
\caption{\textbf{Real-world experimental setup.} We consider four tasks, namely Pick-and-Place, Insert-Hexagon-Block, Insert-Triple-Column-Block, and Erase-Whiteboard, covering multi-step, precise, and contact-rich manipulation. The red and blue areas denote the in-distribution (ID) and out-of-distribution (OOD) evaluation regions, respectively.}
\vspace{-0.3cm}
\label{fig:task_details}
\end{figure*}

\subsection{Digital Twin RL Warm Up}
Although Stage~I expands the spatial coverage of SFT data, directly launching online RL on real robots remains non-trivial due to a bottleneck that spatial coverage alone cannot resolve: the \emph{trajectory style mismatch} between $\mathcal{D}_{\mathrm{sft}}$ and RL interaction data. $\mathcal{D}_{\mathrm{sft}}$ consists of passive, goal-directed demonstrations, whereas online RL requires interactive, trial-and-error trajectories for reliable value estimation. As a result, the replay buffer at the onset of real-world RL is dominated by passive expert-style data, leading to unstable value estimation and poor early-stage exploration.
To address this mismatch, we introduce a twin RL warm-up stage that performs parallel RL rollouts in the digital twin to generate RL-style interaction data, providing a better initialization for the replay buffer and stabilizing subsequent on-robot learning.

\noindent \textbf{Twin Online RL Stage.}
As shown in Fig.~\ref{fig:method} Stage II, we perform parallel online RL in the digital twin. In this stage, the policy $\pi_{\psi}$ is initialized from the SFT model and trained through interaction with $N$ parallel twin environments. 
A joint objective is adopted that combines RL with an imitation-based regularization term to stabilize policy updates during twin-based online interaction.
The objective is defined as:
\begin{equation}
\mathcal{L}_{\pi}^{twin}(\psi) = \beta \mathcal{L}_{\pi}^{IL} + \eta \mathcal{L}_{\pi}^{RL},
\end{equation}
where $\mathcal{L}_{\pi}^{\text{IL}}$ is the same loss used in SFT, and $\mathcal{L}_{\pi}^{RL}$ is the RL objective defined as
$\mathcal{L}_{\pi}^{RL} = -\mathbb{E}_{s \sim \mathcal{D}, a \sim \pi_{\psi}(\cdot|s)} [Q_{\theta}(s, a)]$
, which encourages actions with higher critic-estimated $Q$ values~\cite{haarnoja2018soft}.
The value function $Q_{\theta}$ is updated via standard temporal-difference learning. Through this process, the twin online RL stage efficiently collects diverse trajectories $\tau_{\text{twin}}$, including successful executions, failures, and recovery behaviors, which are stored in the twin replay buffer $\mathcal{D}_{\text{twin}}$.
Note that parallel processing enables us to generate a set of rollouts in about 1 minute (e.g., 30 steps per episode).
Due to the distribution gap between demonstration data and RL-style interaction data, early online learning may exhibit instability. 
Therefore, after efficient online learning in the digital twin, the real-world replay buffer is initialized with data transferred from the twin buffer $\mathcal{D}_{real}^{init} \leftarrow \mathcal{D}_{twin}$.
Twin RL warm-up shifts the replay buffer toward a style-compatible initialization that provides more balanced training signals, thereby reducing instability and mitigating performance degradation during the transition from offline SFT to real-world online RL. Moreover, this strategy helps prevent catastrophic forgetting in configurations that already exhibit strong performance during subsequent targeted HiL-guided online RL.

\subsection{Sim-to-Real Guided Real-World RL}
While twin RL offers a strong initialization, real-world physical interaction remains crucial for precise manipulation and contact-rich tasks. We therefore leverage twin RL rollouts to guide real-world online RL, enabling targeted exploration and more effective use of physical interaction.
As shown in Fig.~\ref{fig:method} Stage III, we leverage the digital twin to identify failure-prone regions of the state space and guide the initial-state distribution for real-world online RL. 
Unlike prior curriculum or reset-based strategies that rely on real-world rollouts~\cite{luo2025precise}, the digital twin enables low-cost and systematic evaluation of policy performance across a wide range of initial configurations without consuming physical interaction budget.
Specifically, we evaluate the current policy in the digital twin and construct a targeted set of initial configurations, $\mathcal{S}_{\text{target}} = \left\{\, s_0 \;\middle|\; SR(s_0) < \tau \,\right\}$, where $SR(s_0)$ denotes the empirical success rate from state $s_0$, and $\tau$ is a proficiency threshold. 
During real-world online interaction, episode resets are prioritized from $\mathcal{S}_{\text{target}}$, enabling the learning process to focus limited physical interaction budget on challenging states.
To further reduce the cost and risk of exploration in challenging regions, we incorporate a HiL mechanism during real-robot training~\cite{luo2025precise}. The resulting intervention trajectories are stored in the replay buffer and used for subsequent policy updates. 
Unlike existing HiL-based approaches, we introduce a novel twin guidance mechanism in which the digital twin continuously informs where to apply HiL intervention during real-world RL, significantly improving sample efficiency and convergence speed.

%% file: section/experiment.tex
\section{Experiment and Results}

We first detail the experimental setup in Section~\ref{sec: exp-setup}, followed by extensive real-world comparisons against baselines in Section~\ref{sec: exp-real}. 
Each component of TwinRL is then ablated in Section~\ref{sec: exp-abla}, and robustness is evaluated in Section~\ref{sec: exp-gen} under varying backgrounds and lighting conditions.
\subsection{Experimental Setup}
\label{sec: exp-setup}

\noindent\textbf{Hardware Platforms.}
For real-world manipulation, we conduct systematic experiments on a 7-DoF Franka Emika Research 3 (FR3) robot.
Our setup features a dual-camera perception system, comprising a fixed third-person view for global context and a wrist-mounted camera for fine-grained details. More details of the setup are provided in~\hyperref[ap:RWS]{Appendix~B}.

\begin{figure*}[t]
\centering
\includegraphics[width=0.98\textwidth]{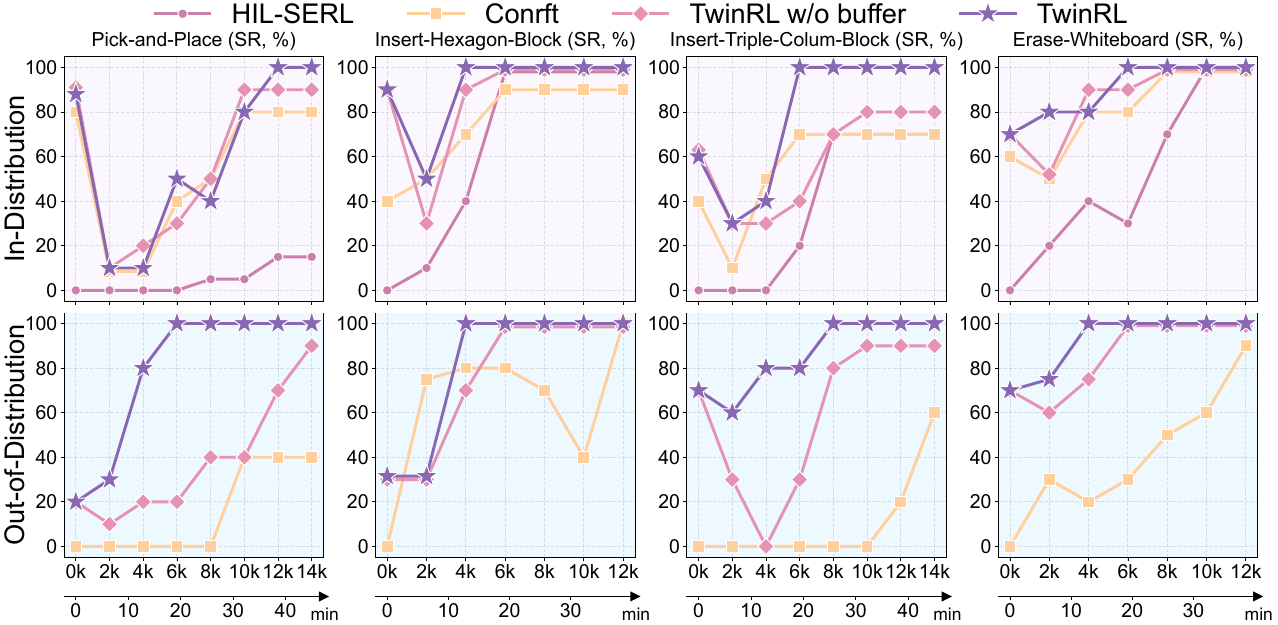}
\caption{\textbf{Real-world Experiments.} We report success-rate curves for online RL across four manipulation tasks under both ID and OOD settings. The y-axis shows success rate, and the x-axis reports both online training time and our model training steps.}
\label{fig:main_exp}
\end{figure*}

\noindent\textbf{Experiment Protocol.}
As shown in Fig.~\ref{fig:task_details}, we discretize each task workspace into a grid, and assign each episode to a grid cell based on the manipulated object’s center location at task completion.
In the figure, the red area denotes (1) in-distribution (ID) regions covered by the collected real-world demonstrations, while the blue area denotes (2) out-of-distribution (OOD) regions that are not covered by real-world data.
Unlike prior work that primarily uses digital twins for data augmentation, our goal is to demonstrate that digital twins can serve as efficient amplifiers and guides for online RL in regions not covered by real-world data.
For time reporting, we define convergence time as the total wall-clock time of real-world online RL (Stage III), measured from the first real-robot rollout until the policy achieves 10 consecutive successful episodes without human intervention.

\subsection{Real-World Experiments}
\label{sec: exp-real}
\noindent\textbf{Implementation Details and Baselines.}
We use Octo-Small~\cite{team2024octo} as a shared VLA backbone across all baselines, with each task defined by a fixed language instruction.
To ensure a fair comparison, we strictly control real-world demonstration data usage across all methods.
\textbf{HiL-SERL}~\cite{luo2025precise} deploys the policy directly onto the physical robot to perform Human-in-the-Loop (HiL) online RL training. \textbf{ConRFT}~\cite{chen2025conrft} serves as a strong baseline utilizing 30 real-world demonstrations from the ID region; it first undergoes the first-stage Cal-ConRFT~\cite{chen2025conrft}, followed by HiL online RL fine-tuning in the real world. 
All loss supervision follows the official paper. \textbf{Our method} uses the same 30 real-world demonstrations and further leverages the digital twin as an exploration amplifier by adding 60 synthetic trajectories in the ID region and 30 synthetic trajectories covering the OOD region. We report two variants: TwinRL w/o buffer and TwinRL. They differ in whether a twin replay buffer, collected via parallel online RL in the digital twin, is used to initialize training before real-world online RL.
During real-world RL, we evaluate each region every 2k steps with at least 10 random rollouts.

\textbf{Metrics and Results.}
We report success rate (SR) as a function of real-world training time and training steps to compare convergence speed and final performance in both ID and OOD regions. Figure~\ref{fig:main_exp} summarizes the online training curves across four tasks.
\textbf{Initialization} (0-step). TwinRL starts with substantially higher SR at the onset of real-world online RL (i.e., before any real interaction). This gain is attributed to the exploration space expansion during SFT, where digital-twin trajectories broaden the support of the pretraining distribution and yield a stronger deployment prior.
\textbf{In ID regions}, the key difference between TwinRL and the baselines lies in whether sim-to-real guided exploration is available. Most TwinRL variants exceed 90\% SR and converge noticeably faster than the baselines, indicating that our digital-twin online rollout strategy serves as an effective guide for real-world RL. By efficiently identifying informative object configurations in the twin, TwinRL focuses HiL interactions on high-value states, accelerating convergence while maintaining high execution accuracy.
\textbf{In OOD regions}, the gap is more pronounced. TwinRL attains high SR with far fewer real-world interactions, whereas ConRFT and HiL-SERL converge more slowly or fail to reach comparable performance under the same interaction budget. This demonstrates that TwinRL effectively expands exploration beyond the spatial support of real demonstrations and enables rapid adaptation to previously unseen configurations.
\textbf{Stability.} All methods exhibit noticeable SR degradation when transitioning from offline SFT to online learning, indicating instability under the distribution shift from SFT to RL-style data. TwinRL minimizes this gap and recovers performance much faster, reaching 100\% SR. This improved stability is consistent with transferring twin RL rollouts to initialize the real replay buffer, which bridges the offline-to-online transition and mitigates early-stage performance collapse.
Finally, we conduct five rounds of evaluation on all four tasks after RL training, each consisting of 10 rollouts in both ID and OOD regions. TwinRL achieves an average SR of 95\% with a standard deviation of 6.71.
Failure analysis is provided in~\hyperref[ap:FCA]{Appendix~E}.
Additional metrics (episode length and intervention rate) and visualizations are provided in~\hyperref[ap:AR]{Appendix~D1\&D2\&D6}. Demo videos are available on the project page.

\begin{table}[t]
\centering
\small
\caption{\textbf{Ablation on exploration space expansion.} We vary the number of twin-generated trajectories added during SFT warm-up and measure the resulting success rate (SR).}
\setlength{\tabcolsep}{2pt}
\renewcommand{\arraystretch}{1.10}
\begin{tabular}{ccccc}
\toprule
\multicolumn{2}{c}{\textbf{Data volume}} &
\textbf{In-Distribution} &
\textbf{Out-of-Distribution} &
\textbf{Avg.} \\
\cmidrule(lr){1-2}\cmidrule(lr){3-3}\cmidrule(lr){4-4}\cmidrule(lr){5-5}
\textbf{In} & \textbf{Out} &
\textbf{SR ($\uparrow \Delta$)} &
\textbf{SR ($\uparrow \Delta$)} &
\textbf{SR ($\uparrow \Delta$)} \\
\midrule
0  & 0  & 40\%  & 0\%   & 27\% \\
30 & 30 & 70\% (+30\%) & 30\% (+30\%) & 57\% (+30\%) \\
60 & 30 & \textbf{80\% (+40\%)} & 40\% (+40\%) & 67\% (+40\%) \\
30 & 60 & 70\% (+30\%) & \textbf{70\% (+70\%)} & \textbf{70\% (+43\%)} \\
\bottomrule
\end{tabular}
\label{tab:twin_aug_ablation}
\end{table}

\begin{table}[t]
\centering
\small
\caption{\textbf{Ablation on the twin buffer.} We report SR and RL training steps. \textbf{S} and \textbf{F} denote successful and failed twin RL trajectories used to populate the replay buffer.}
\setlength{\tabcolsep}{6pt}
\begin{tabular}{l l c c}
\toprule
\textbf{Ablation} & \textbf{Setting} & \textbf{Online Steps} & \textbf{Success Rate} \\
\midrule
\multirow{4}{*}{Twin Buffer}
 & 0 S / 0 F  & 5.0k & 90\% \\
 & 20 S / 0 F  & \textbf{3.5k} & \textbf{100\%} \\
 & 30 S / 0 F  & 4.0k & 90\% \\
 & 20 S / 20 F & 4.5k & 90\% \\
 & 20 S / 40 F & 7.0k & 70\% \\
\bottomrule
\end{tabular}
\label{tab:ablation_twin}
\vspace{-0.1cm}
\end{table}

\begin{figure}[t]
\centering
\includegraphics[width=0.97\columnwidth]{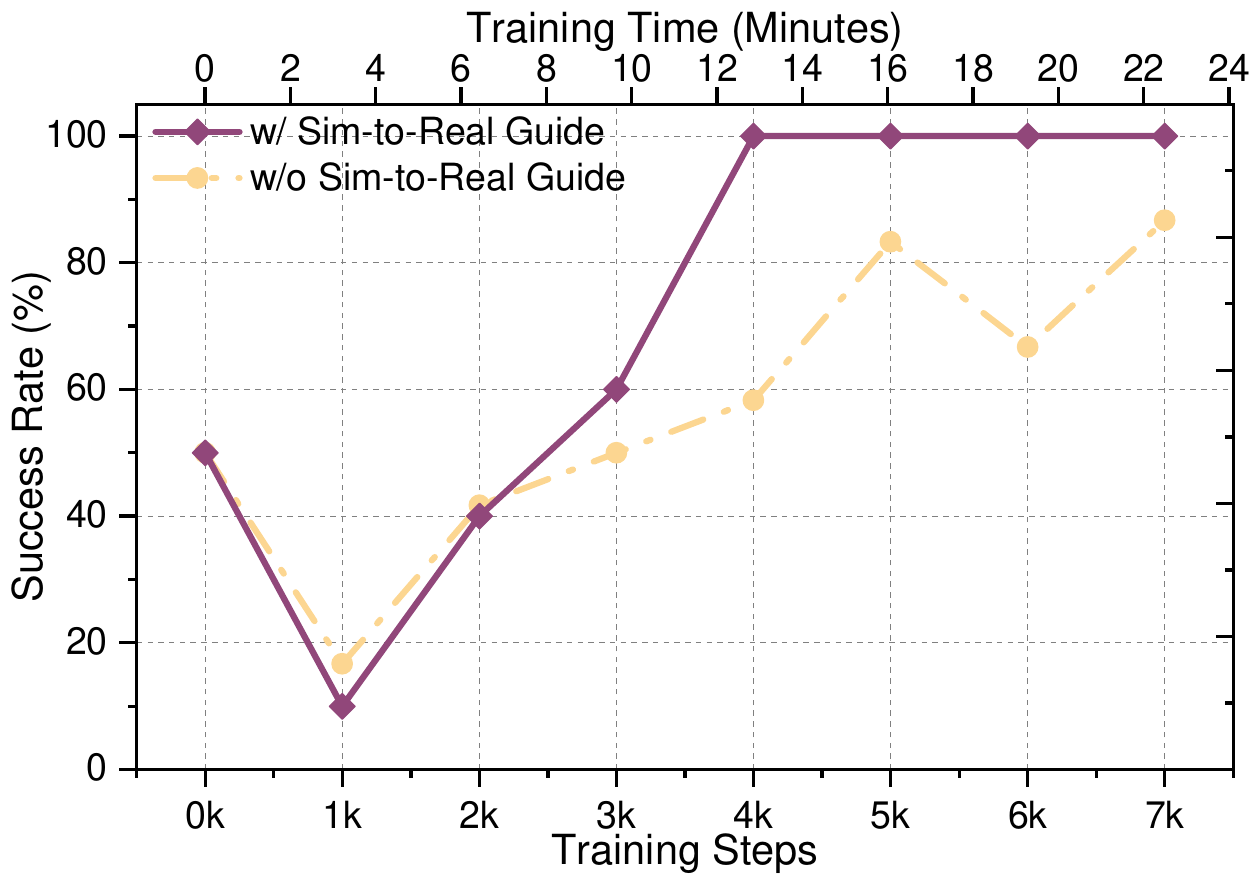}
\caption{\textbf{Ablation on Sim-to-Real-Guided HiL.} The guidance markedly accelerates RL learning, reaching $100\%$ success at around $4$k steps ($\sim$14 min).}
\vspace{-0.2cm}
\label{fig:ablation_hil}
\end{figure}

\subsection{Ablation Study}
\label{sec: exp-abla}
We select Insert-Hexagon-Block for ablation. As shown in Fig.~\ref{fig:task_details}, the evaluation setup is consistent with the main experiments.

\noindent\textbf{Impact of Exploration Space Expansion.}
We analyze how the scale and distribution of twin data affect warm-up performance. Table~\ref{tab:twin_aug_ablation} reports success rates under different ID/OOD synthetic-trajectory augmentations. Compared to the base model, a balanced setting with 30 ID and 30 OOD twin trajectories achieves 57.0\% SR (+30\%), indicating that our digital-twin pipeline can generate high-quality trajectories across the workspace even for the precision-critical task. Increasing the amount of twin data further improves performance: doubling ID data (60/30) yields the largest gain, reaching a peak ID SR of \textbf{80\%}, while increasing OOD data (30/60) improves SR to 70\%. Overall, these results not only validate the quality of data generated by the digital twin but also demonstrate that our exploration space expansion strategy effectively broadens coverage.
We analyze how the scale and distribution of twin data affect warm-up performance. Table~\ref{tab:twin_aug_ablation} reports success rates under different ID/OOD synthetic-trajectory augmentations. Compared to the base model, a balanced setting with 30 ID and 30 OOD twin trajectories achieves 57.0\% SR (+30\%), indicating that our digital-twin pipeline can generate high-quality trajectories across the workspace even for the precision-critical task. Increasing the amount of twin data further improves performance: doubling ID data (60/30) yields the largest gain, reaching a peak ID SR of \textbf{80\%}, while increasing OOD data (30/60) improves SR to 70\%. Overall, these results validate both the quality of twin-generated data and the effectiveness of our method in broadening spatial exploration coverage.
More synthetic data can further help, but it also increases SFT time, creating an accuracy–efficiency trade-off.

\noindent\textbf{Impact of Twin RL Replay Buffer.}
As shown in Fig.~\ref{fig:main_exp}, we have demonstrated the importance of the twin RL replay buffer in stabilizing early-stage learning. Here, we further investigate how the composition of the replay buffer affects the efficiency and stability of online RL. Specifically, we vary the ratio of successful to failed twin rollouts stored in the buffer, ranging from success-only buffers to buffers with an increasing number of failure examples. As shown in Table~\ref{tab:ablation_twin}, most settings ultimately reach over 90\% success in both ID and OOD regions. These results show that seeding the twin buffer with successful trajectories can substantially accelerate online RL training. However, adding more failure rollouts reduces online RL efficiency, contrary to our expectation. 
We attribute this to the fact that failure trajectories sampled from the digital twin often reflect random failure modes rather than meaningful, task-relevant failures encountered during real-world HiL interaction, and thus do not effectively teach the policy to avoid failure states.
In future work, we will explore more principled ways to incorporate informative failures before online RL.

\noindent\textbf{Efficiency of Sim-to-Real-Guided HiL.}
Finally, we evaluate the role of digital-twin-guided HiL during real-world online RL. As shown in Fig.~\ref{fig:ablation_hil}, we compare TwinRL with and without the proposed guidance mechanism, which uses digital-twin rollouts to identify informative initial object configurations and trigger HiL intervention when necessary. The results show that twin-guided intervention significantly reduces real-world training steps and improves convergence speed and success rate. Without guidance, adaptation becomes slower and less sample-efficient, despite identical initialization and replay buffers. These findings highlight that digital twins in TwinRL not only expand exploration support prior to deployment but also actively guide online learning toward informative regions of the state space.
Additional ablation studies are provided in~\hyperref[ap:AR]{Appendix~D3\&D4}, including analyses of how the distribution of twin data affects SFT performance and whether digital twin evaluation is consistent with real-world performance.

\begin{figure}[t]
\centering
\includegraphics[width=0.97\columnwidth]{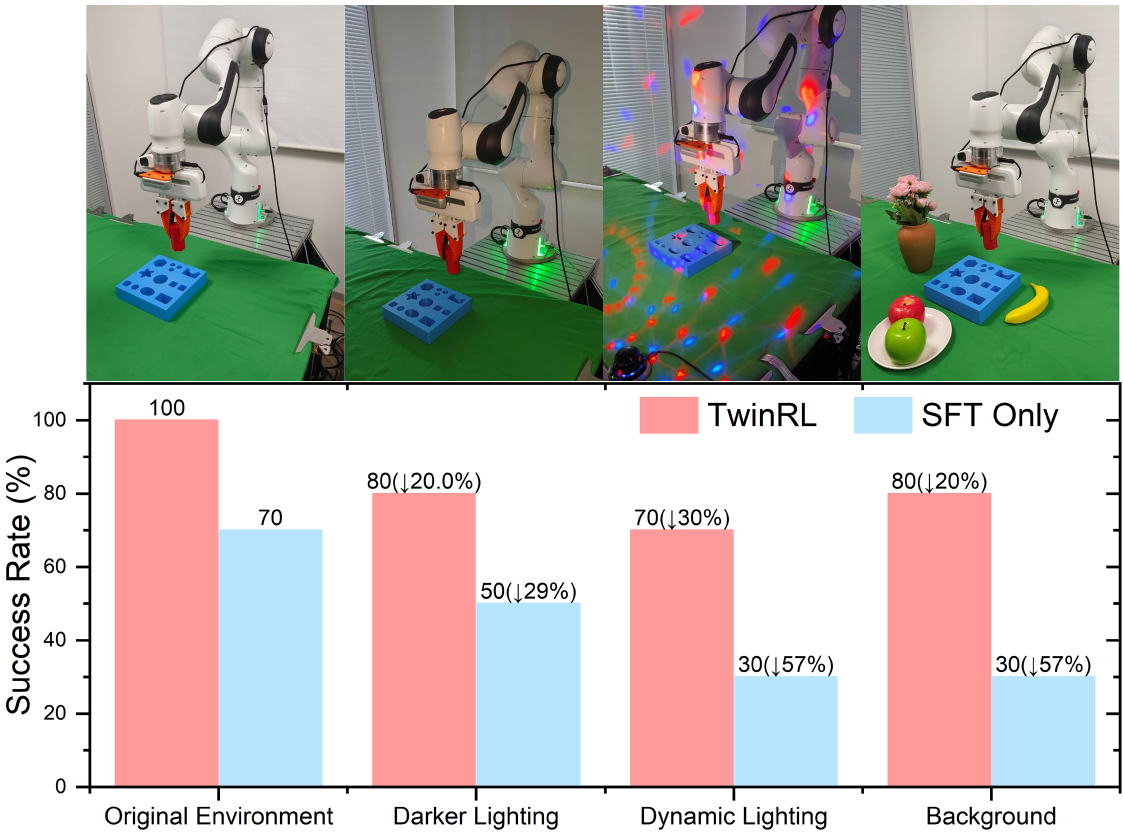}
\caption{\textbf{Robustness.} We compare the SFT policy and the TwinRL-guided online RL policy under previously unseen environmental perturbations; the top row shows examples.}
\vspace{-0.3cm}
\label{tab:generalization_study}
\end{figure}

\subsection{Robustness Analysis}
\label{sec: exp-gen}
We compare the SFT policy and the TwinRL-guided online RL policy under a zero-shot robustness evaluation with previously unseen environmental perturbations, including background clutter and lighting variations. As shown in Fig.~\ref{tab:generalization_study}, we consider three test conditions: Background (adding task-irrelevant objects to create clutter), Darker lighting (uniform illumination changes), and Dynamic lighting (dynamic colored lighting with moving light patterns). All perturbations are applied to the hexagonal block insertion task. The results show that, despite the observation distribution shift, TwinRL suffers only a minor performance drop, whereas the SFT-only model exhibits a substantially larger degradation.
The results highlight the importance of real-world RL (Stage III), which pushes the policy toward more stable control and noise-tolerant decision boundaries in the physical environment. They also demonstrate the effectiveness of the TwinRL system, which further improves robustness by using twin-guided HiL to focus real-world interactions on high-information configurations.
Additional robustness results on the Pick-and-Place task, reported in~\hyperref[ap:AR]{Appendix~D5}.